\documentclass[10pt,twocolumn,letterpaper]{article}

\usepackage{iccv}
\usepackage{times}
\usepackage{epsfig}
\usepackage{graphicx}
\usepackage{amsmath}
\usepackage{amssymb}
\usepackage{xspace}
\usepackage{algorithm}
\usepackage[algo2e]{algorithm2e} 
\usepackage{algorithmic}
\usepackage[mathlines,switch]{lineno}
\usepackage{longtable}
\usepackage{makecell}

\usepackage{array}
\usepackage{multirow}
\usepackage{setspace}
\usepackage[table,xcdraw]{xcolor}
\usepackage{booktabs}
\usepackage{bbm}
\usepackage{authblk}
\usepackage{pifont}
\newcommand\extrafootertext[1]{%
    \bgroup
    \renewcommand\thefootnote{\fnsymbol{footnote}}%
    \renewcommand\thempfootnote{\fnsymbol{mpfootnote}}%
    \footnotetext[0]{#1}%
    \egroup
}

\usepackage[pagebackref=true,breaklinks=true,letterpaper=true,colorlinks,bookmarks=false]{hyperref}

\iccvfinalcopy 

\def\iccvPaperID{12618} 

\ificcvfinal\fi

\begin{document}

\title{UniFormaly: Towards Task-Agnostic Unified Framework \\ for Visual Anomaly Detection}
\author{Yujin Lee\qquad Harin Lim\qquad Seoyoon Jang\qquad Hyunsoo Yoon\\
Department of Industrial Engineering, Yonsei University\\
Seoul, South Korea\\
{\tt\small \{yjlee9040,harini1029,tjdbs70002,hs.yoon\}@yonsei.ac.kr}}

\maketitle
\ificcvfinal\thispagestyle{empty}\fi

\begin{abstract}
    Visual anomaly detection aims to learn normality from normal images, but existing approaches are fragmented across various tasks: defect detection, semantic anomaly detection, multi-class anomaly detection, and anomaly clustering. This one-task-one-model approach is resource-intensive and incurs high maintenance costs as the number of tasks increases. We present \textit{UniFormaly}, a universal and powerful anomaly detection framework. We emphasize the necessity of our off-the-shelf approach by pointing out a suboptimal issue in online encoder-based methods. We introduce Back Patch Masking (BPM) and top $k$-ratio feature matching to achieve unified anomaly detection. BPM eliminates irrelevant background regions using a self-attention map from self-supervised ViTs. This operates in a task-agnostic manner and alleviates memory storage consumption, scaling to tasks with large-scale datasets. Top $k$-ratio feature matching unifies anomaly levels and tasks by casting anomaly scoring into multiple instance learning. Finally, \textit{UniFormaly} achieves outstanding results on various tasks and datasets. Codes are available at \href{https://github.com/YoojLee/Uniformaly}{https://github.com/YoojLee/Uniformaly}.
\end{abstract}


\section{Introduction}

\begin{figure}[ht!]
\begin{center}
\includegraphics[width=1.0\linewidth]{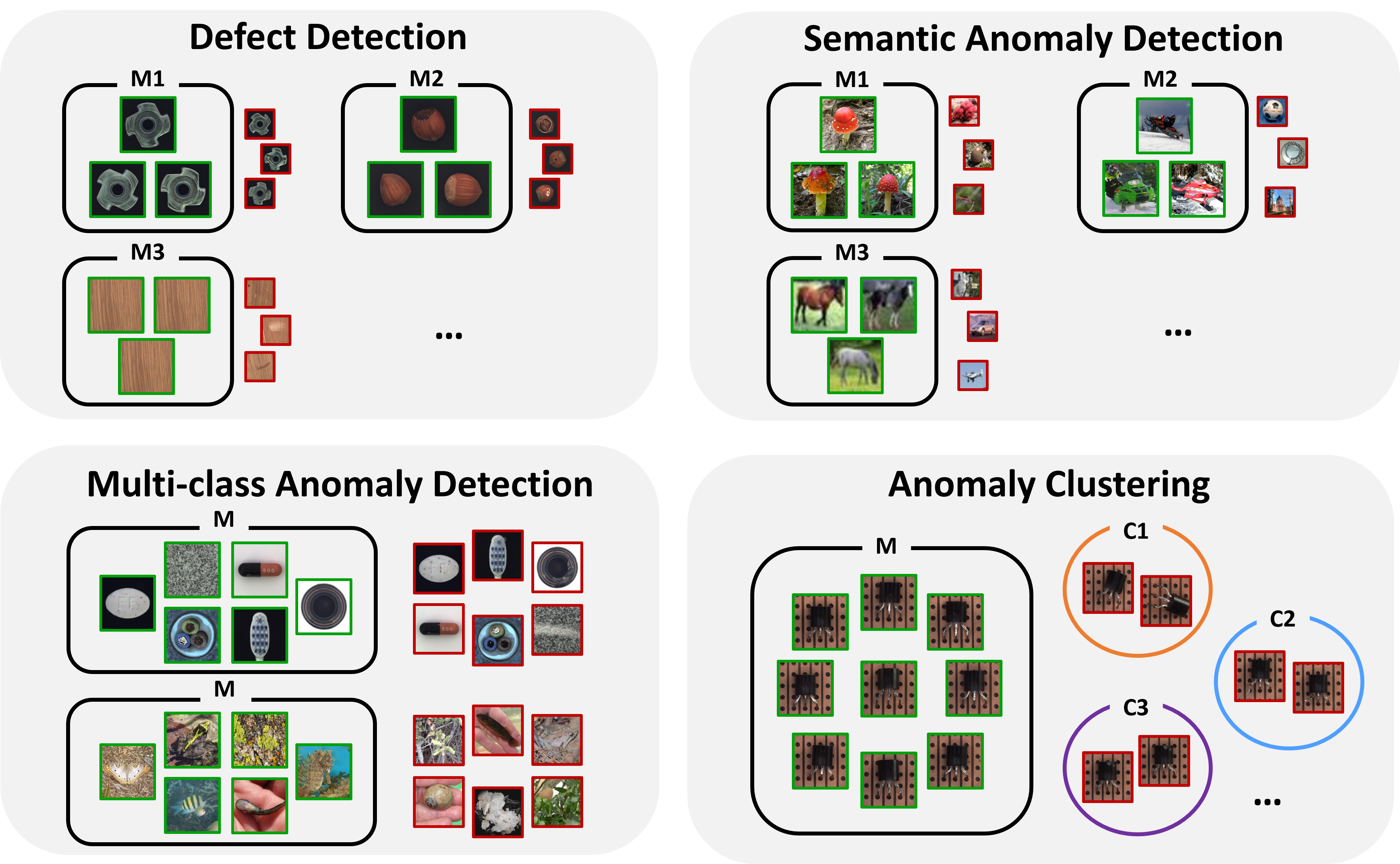}
\end{center}
    \caption{\textbf{Examples of different anomaly detection tasks.} \textit{UniFormaly} successfully detects 1) local defects such as scratches, 2) semantic anomalies of different classes, 3) anomalies involving multiple classes simultaneously, and 4) clusters of anomaly types, such as cuts or holes.}
\label{fig:onecol}
\end{figure}

    Anomaly detection, often referred to as novelty detection, is a well-established field in computer vision, dedicated to identifying anomalous data that deviate from normality. With the exponential growth of image data, the demand for robust visual anomaly detection systems spans a wide spectrum of industries. Consequently, various tasks have emerged within the field of anomaly detection. These tasks primarily encompass defect detection and semantic anomaly detection, which share a common objective but differ in the level of anomalies. In defect detection (also known as industrial anomaly detection), as illustrated in Fig.~\ref{fig:onecol}, methods seek to identify anomalies that are locally deviant from the normal appearance while being semantically identical. On the other hand, semantic anomaly detection targets semantically different anomalies that do not belong to any known normal class. Semantic anomaly detection can be further utilized in autonomous driving. Moreover, recent advancements have introduced multi-class anomaly detection, where a single framework processes multiple classes or datasets as normal for anomaly detection~\cite{you2022a}. In addition to the primary task of differentiating anomalies from normal data, there has been a growing effort to comprehensively understand and classify various types or categories of anomalies~\cite{sohn2023anomaly}.
   
\begin{figure*}[t]
\begin{center}
    \includegraphics[width=0.9\linewidth]{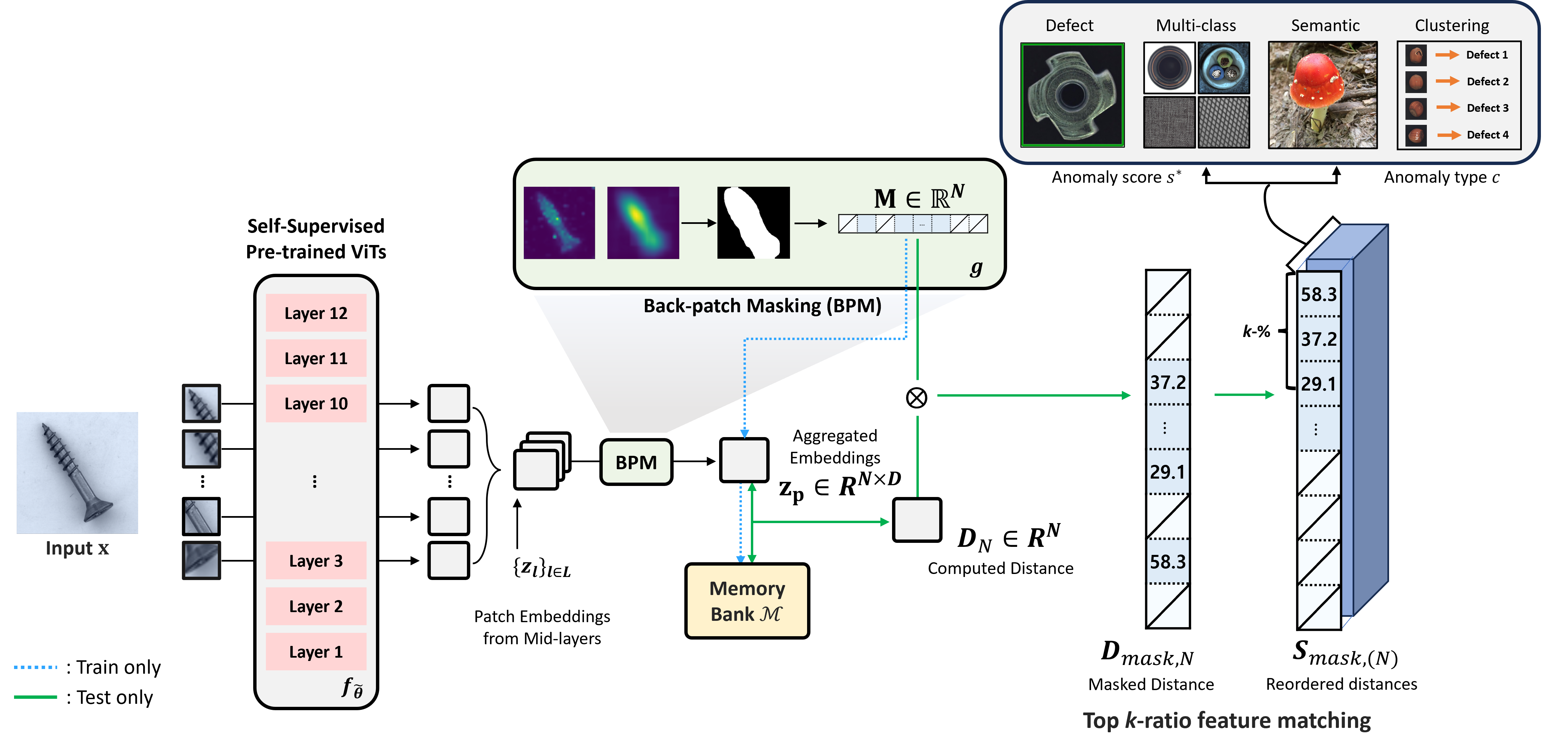}
\end{center}
    \caption{\textbf{Overview of \textit{UniFormaly}.} Normal patch embeddings extracted from self-supervised pre-trained ViTs are aggregated and fed into a memory bank. Given novel data, we extract patch features and compute the nearest distance. To prevent \textit{UniFormaly} from attending to unrelated regions, we apply Back Patch Masking and reorder scores. Finally, we compute the anomaly score $s^*$ and anomaly type $c$ based on the top $k$-ratio of reordered patches.}
\label{fig:architecture}
\end{figure*}

    Meanwhile, in the recent AI community, a unified system has been extensively developed in various domains due to its ability to avoid task-specific design and training, reducing memory consumption and maintenance costs. However, previous anomaly detection approaches remain fragmented across different tasks, with limited exploration of unification. This fragmentation is primarily driven by the prevailing emphasis on achieving high accuracy within specific tasks, which is considered more significant than in other domains. Nevertheless, since anomaly detection tasks all aim to learn normality, adopting the conventional approach of one-task-one-model, which necessitates task-specific training, may be less efficient and more resource-intensive. Therefore, the anomaly detection domain still requires a unified and highly effective system. In this work, we propose \textit{UniFormaly}, a powerful and universal anomaly detection system, encompassing various anomaly detection tasks across anomaly levels (defect, semantic), objectives (binary detection, localization, type clustering) even in a wide range of normality boundaries (i.e., multi-class anomaly detection).
        
    In the anomaly detection literature, approaches based on online normality learning, such as reconstruction or self-supervised learning, have been prevalent in the anomaly detection literature, while pre-trained encoder-based methods have recently emerged as highly effective anomaly detection systems. However, these online approaches require proxy tasks that may not align with the target task (i.e., anomaly detection) and require prior knowledge of anomalies in the datasets. In contrast, pre-trained encoders lack proxy tasks, making them a promising foundation for unified anomaly detection. In Section \ref{sec:representation_analysis}, we empirically highlight the superiority of pre-trained encoder methods as the foundation of a unified system compared to other approaches. In addition, through comparative studies across different encoder architectures and levels of supervision, we confirm that self-supervised representations from Vision Transformers (ViTs) hold great potential for achieving unified anomaly detection.
        
    We introduce two critical components, Back Patch Masking (BPM) and top $k$-ratio feature matching, to construct an effective and unified anomaly detection system. While pre-trained encoder-based methods hold significant potential for unifying anomaly detection, they face challenges related to memory storage consumption due to the need for a memory bank containing normal features. In semantic anomaly detection, with a substantial number of normal images, memory storage becomes a prominent concern, especially when dealing with multiple normal classes. A previous study~\cite{roth2022towards} has tried to reduce the size of the memory bank using coreset subsampling with minimal impact on its performance. Building upon this prior work, we introduce Back Patch Masking (BPM) to further reduce the size of the memory bank. BPM drops the less informative background patches with the employment of fixed self-attention to guide the system in non-target rejection. Since the self-attention is derived from a fixed pre-trained encoder, our BPM operates in a task- and dataset-agnostic manner. Additionally, BPM enables our system to handle both false-positive and false-negative cases, contributing to its high effectiveness.
    
    Top $k$-ratio feature matching, another pivotal component, contributes to its generality and effectiveness. In the current literature, the level of representation in pre-trained encoder approaches varies across tasks. For example, defect detection utilizes mid-level representation for effective defect detection~\cite{roth2022towards}, while semantic anomaly detection relies on global representation~\cite{sun2022out}. To achieve unification, we adopt mid-level features and employ them in a multiple-instance learning fashion. Precisely, we extract the top $k$\% anomalous patch features and utilize them across various anomaly detection tasks. Furthermore, we demonstrate that this top $k$-ratio feature matching enables anomaly clustering, marking a significant step toward developing a general-purpose anomaly detection system. In Section \ref{sec:uniformaly_method}, we provide an in-depth explanation for BPM and top $k$-ratio feature matching.
    
    Through extensive experiments, we demonstrate the versatility of \textit{UniFormaly} across a spectrum of tasks: (1) defect detection (MVTecAD, BTAD, MTD, CPD), (2) semantic anomaly detection (CIFAR-10, CIFAR-100, ImageNet-30, Species-60), (3) multi-class anomaly detection (MVTecAD, CIFAR-10, Species-60), (4) anomaly clustering (MVTecAD), and (5) low-shot scenarios (MVTecAD). Our results are remarkable, demonstrating state-of-the-art or competitive performance, such as an AUROC of 99.32 on MVTecAD and 97.6 on ImageNet-30. We also conduct ablation studies on BPM and top $k$-ratio feature matching, confirming their substantial contributions to the generality and effectiveness of our anomaly detection system.

\section{Related Works}

Over the past decade, deep learning approaches, such as reconstruction or self-supervised learning, have gained prominence in addressing various anomaly detection tasks. Reconstruction-based approaches~\cite{zavrtanik2021reconstruction,you2022a} assume that anomalies cannot be accurately reconstructed, while self-supervised learning-based approaches rely on augmentations of normal images~\cite{zhang2022deep, tack2020csi, li2021cutpaste}. Specifically, ~\cite{zavrtanik2021reconstruction} combines elements of both by incorporating the inpainting task within an autoencoder.

While the two approaches focus on learning normal representations from scratch, recent methods leverage rich representations from encoders pre-trained on large-scale image datasets. Several works propose to adapt pre-trained features for anomaly detection~\cite{reiss2021panda, yang2022learning}. In ~\cite{reiss2021panda}, feature adaptation is achieved through the use of compactness loss and joint optimization, and ~\cite{yang2022learning} leverages an additional feature estimation network to maximize feature correspondence. Other works use off-the-shelf representations and feed them into a normal feature memory bank to determine anomalies~\cite{cohen2020sub,defard2021padim,roth2022towards}. In this line of work, it is important to make the memory bank compact but highly representative. ~\cite{cohen2020sub} uses image-level representations for anomaly detection and reduces the size of the memory bank by reusing them for anomaly localization. ~\cite{defard2021padim} computes statistical estimates for normal features, and ~\cite{roth2022towards} summarizes the memory bank with coreset subsampling. These approaches based on off-the-shelf representations have gained popularity due to their simplicity and effectiveness.

However, despite their effectiveness, previous approaches in the literature have been developed in a fragmented manner with a lack of attempts for unification. While ~\cite{you2022a} unifies models across classes into a single architecture, it does not address grouping anomalies or detecting anomalies in high-resolution natural images. In contrast, our method achieves unified detection at various anomaly levels and enables clustering of anomalies arising in subtle regions. Furthermore, our method handles low normal-shot anomaly detection, making it more practical for real-world unified anomaly detection systems.

In pre-trained encoder-based methods, it is critical to generate effective feature maps for anomaly detection. This issue encompasses two factors: encoder architecture and feature aggregation. While supervised ConvNets are widely used in pre-trained encoder-based methods, we explore alternative representation choices for anomaly detection. As for feature aggregation, previous pre-trained encoder methods often use a feature map from the last layer without aggregation~\cite{bergman2020deep} or employ rescaling and subsequent concatenation~\cite{cohen2020sub} for aggregation. However, more sophisticated feature aggregation techniques are required for defect detection. Recent approach~\cite{roth2022towards} has introduced more complex feature aggregation methods that leverage local neighborhoods with a fixed window size and apply adaptive pooling in the channel dimension. Despite the significant improvement in defect detection performance, it is complex and sensitive to hyperparameters such as neighborhood size and pooling dimension, limiting scalability across various tasks or datasets. In contrast, \textit{UniFormaly} simplifies feature aggregation with minimal hyperparameters, leading to robustness across datasets and tasks.

Patch-level anomaly learning has gained prominence in image anomaly detection, where images are treated as collections of patches. This method computes anomaly scores for individual patches, typically through explicit feature training. For instance, ~\cite{yi2020patch} employs Support Vector Data Description (SVDD) for patch-level anomaly detection, and ~\cite{ahn2022application} extends this with mini-batch K-means clustering and metric learning. Moreover, ~\cite{zhang2022pedenet} incorporates density estimation using the Gaussian Mixture Model (GMM) for clustering.

In contrast, some approaches ~\cite{defard2021padim,roth2022towards} integrate patch-level representations with offline encoders, utilizing nearest neighbor retrieval from a memory bank of normal features. However, these offline encoder-based methods primarily focus on subtle defect detection and do not address semantic anomaly detection, which involves identifying higher-level anomalies. To bridge this gap, our approach extends patch-level anomaly learning with Back Patch Masking and top-k ratio feature matching, enabling task-agnostic anomaly detection and the identification of higher-level anomalies.

\section{Rethinking Off-the-shelf Representations for Unified Anomaly Detection}\label{sec:representation_analysis}

\begin{figure}
    \begin{center}
        \includegraphics[width=0.8\linewidth]{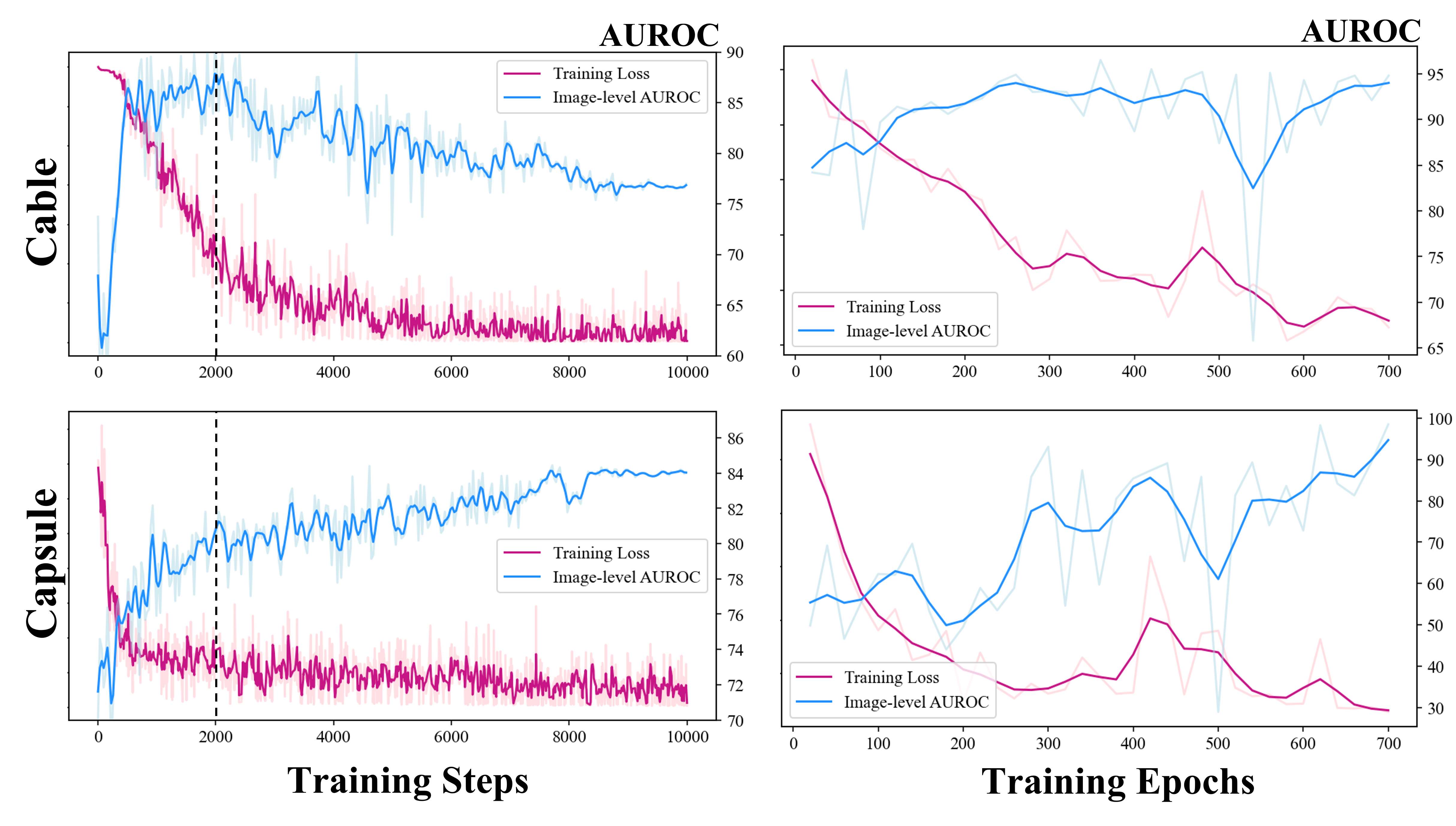}
    \end{center}
        \caption{\textbf{Suboptimal issues with performance fluctuation.} We present the training and monitoring curves of the self-supervised learning-based method~\cite{li2021cutpaste} (left) and the reconstruction-based method~\cite{zavrtanik2021draem} (right) on two classes of MVTecAD.}
    \label{fig:periodic_fluctuation}
    \end{figure}

Online anomaly detection methods, such as reconstruction or self-supervised learning methods, often require explicit learning of normality using proxy tasks. However, the discrepancy between proxy tasks and target tasks renders online methods less suitable for general-purpose anomaly detection. Fig.~\ref{fig:periodic_fluctuation} presents this disagreement by showcasing the training curve of proxy tasks from ~\cite{zavrtanik2021draem,li2021cutpaste} alongside periodic evaluation results. While the training loss shows a decreasing trend, the AUROC does not consistently follow the same pattern. Moreover, the optimal points vary across different object categories. Since ground truth labels are unavailable for anomaly detection when transferring to other datasets and tasks, periodic evaluation for monitoring performance become infeasible. As a result, the reliability of online methods is compromised, making it impractical to build a unified system based on these approaches.

We advocate for a pre-trained encoder-based approach that relies on rich representations learned from large-scale datasets, eliminating the need for proxy task learning on the specific dataset. The choice of representation is pivotal in these methods. While supervised ConvNets have traditionally been used for anomaly detection, they may not be optimal for constructing a unified system applicable to diverse datasets due to potential constraints from label information from the pre-training dataset. Moreover, conventional ConvNets with fixed kernel sizes lack flexibility and necessitate additional aggregation steps, increasing the involvement of hyperparameters. This hinders the development of a task-agnostic anomaly detection system. To validate these arguments, we conduct anomaly detection experiments considering supervision levels and architectural aspects, utilizing heterogeneous datasets (i.e., MVTecAD) from the pre-training datasets (i.e., ImageNet) for evaluation. This assessment focuses solely on measuring representational power, excluding factors such as aggregation.

\begin{figure}
    \begin{center}
        \includegraphics[width=0.8\linewidth]{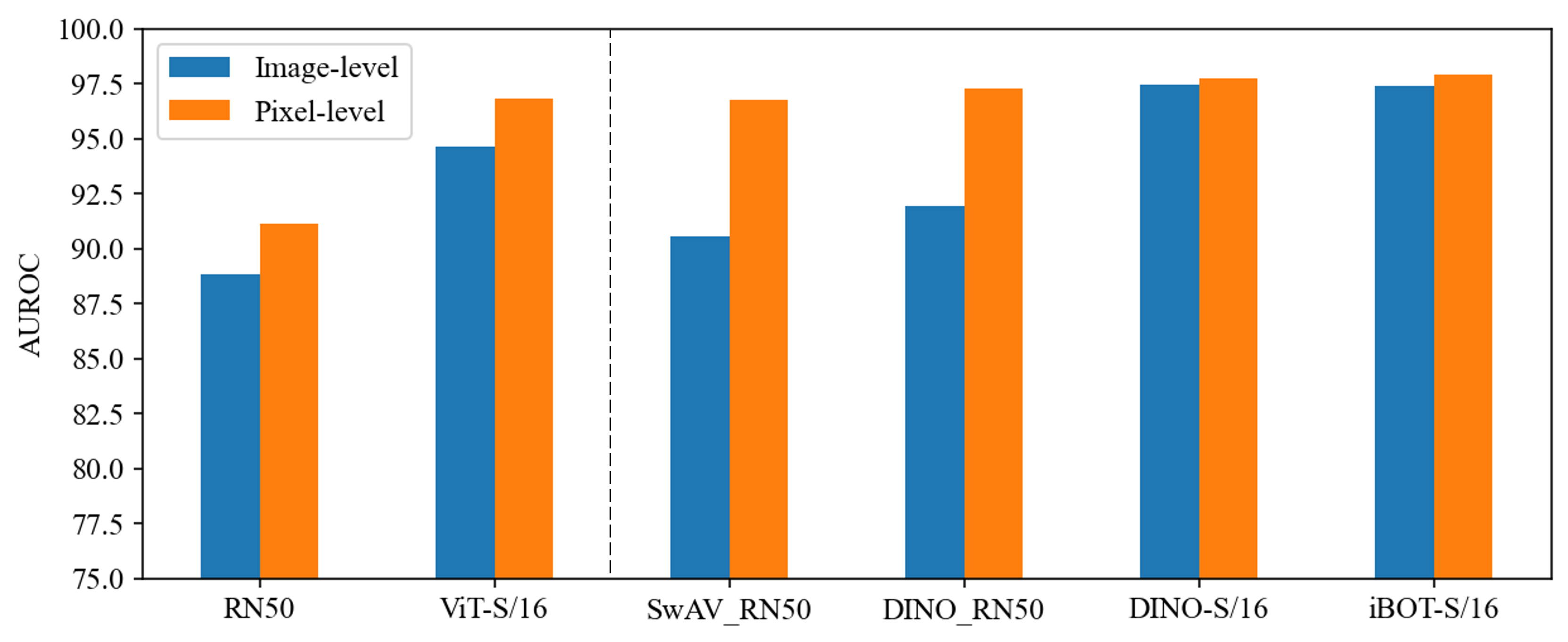}
    \end{center}
        \caption{\textbf{Comparison of different architectures and levels of supervision.} Comparison among supervised (left side of the dotted line) and self-supervised~\cite{caron2020unsupervised, caron2021emerging, zhou2021ibot} (right side of the dotted line) representation from ResNet50~\cite{he2016deep} and ViT-S/16~\cite{dosovitskiy2020image}, with the approximate number of parameters.}
    \label{fig:compare_sup_selfsup}
    \end{figure}

Fig.~\ref{fig:compare_sup_selfsup} illustrates the results, demonstrating that supervised ResNet, without specific aggregation, fails to achieve high performance based solely on representation. In contrast, self-supervised ViT demonstrates strong performance in both image-level and pixel-level AUROC, with minimal variation between the two. While supervised ConvNets have shown good performance in defect detection, particularly when combined with effective aggregation~\cite{roth2022towards}, they may not necessarily be the optimal choice in terms of sole representation. In fact, self-supervised ViT, which has achieved impressive results in image classification, exhibits effectiveness across different industrial image benchmarks, further highlighting its potential as a scalable representation for task-agnostic anomaly detection.

\section{\textbf{\textit{UniFormaly}}: Towards Task-Agnostic Unified
Framework for Visual Anomaly Detection}\label{sec:uniformaly_method}
\textit{UniFormaly} achieves unified anomaly detection based on a patch-level memory bank with the assistance of Back Patch Masking (BPM) (Section \ref{sec:bpm}) and Top $k$-ratio feature matching (Section \ref{sec:tkfm}). An overview of \textit{UniFormaly} is provided in Figure \ref{fig:architecture} and Algorithm \ref{alg:UniFormaly}.
\begin{algorithm}[ht!]
\modulolinenumbers[1]
\caption{Unified anomaly detection with \textit{UniFormaly}}\label{alg:UniFormaly}
\SetAlgoLined
\SetKwInput{KwInput}{Input}
\SetKwInOut{Output}{Output}

\KwInput{Normal training data $D_{train}$, Test data $D_{test}$, self-supervised ViT $f$, a set of layer indices $\{l_1,...,l_L\}$, smoothing kernel $g$, kernel size $n$, k-ratio $k$, clustering function $C$}

\BlankLine
\textbf{Training Stage:} \\
Initialize memory bank $\mathcal{M}$ with an empty set. \\
\ForEach{$d_i$ $\in$ $D_{train}$}{
    \begin{algorithmic}[1]
    \STATE $\mathcal{P}_i \leftarrow \frac{1}{L}\sum_{j=1}^{L}f_{l_j}(d_i)$  \\
/* Start BPM */ 
    \STATE Get self-attention $S_A$ from $f_{l_L}$
    \STATE $S_A \leftarrow g(n) \circ \!\, S_A$ \ \ // update $S_A$ using $g$
    \STATE $\mathrm{M} \leftarrow$ Binarize $S_A$ with a threshold $\lambda$
    \STATE $\mathcal{P}_{mask, i} = \mathrm{M} \otimes \!\, \mathcal{P}_i$  \ \ // cancel out unrelated patches\\
/* End BPM */
    \STATE $\mathcal{M}$ ← $\mathcal{P}_{mask, i}$
    \end{algorithmic}
}
Update a normal memory bank $\mathcal{M}_s$ by sampling

\BlankLine
\textbf{Testing Stage:} \\
\ForEach{$d_i^{test}$ $\in$ $D_{test}$}{
    \begin{algorithmic}[1]
    \STATE Repeat the step 1-5 in the training stage to get $\mathcal{P}^{test}_{mask, i}$\;
    \STATE Compute the distance $\mathcal{D}_N$ from $\mathcal{M}_s$ to each patch in $\mathcal{P}^{test}_{mask, i}$\;
    \STATE Take set of patches $\mathcal{P}_k^{test}$ with top $k$\% of $\mathcal{D}_N$
    \STATE anomaly score $s^* \leftarrow \sum \mathcal{D}_{\mathcal{P}_k^{test}}$\;
    \STATE anomaly type $c \leftarrow C(\mathcal{P}_k^{test}) $\;
\end{algorithmic}
}

\BlankLine
\Output{$s^*, c$}
\end{algorithm}
\subsection{Back Patch Masking}\label{sec:bpm}
Since pre-trained encoder approaches do not require explicit learning of normality from normal images, the model can be susceptible to distractions from spurious signals in the background. Previous research has explored training an attention module alongside the detection model to mitigate its susceptibility to background distractions~\cite{zhang2020multi, venkataramanan2020attention, wu2021learning}. However, in the case of \textit{UniFormaly}, where normality is not explicitly learned, it is unreasonable to train instance-aware attention separately. Furthermore, the attention map tuned for normality is not guaranteed to effectively generalize across diverse datasets and tasks due to the inherent disparities between proxy tasks and the target task. These limitations make traditional attention-based approaches unsuitable for constructing a unified anomaly detection system.

\begin{figure}[ht!]
    \begin{center}
        \includegraphics[width=\linewidth]{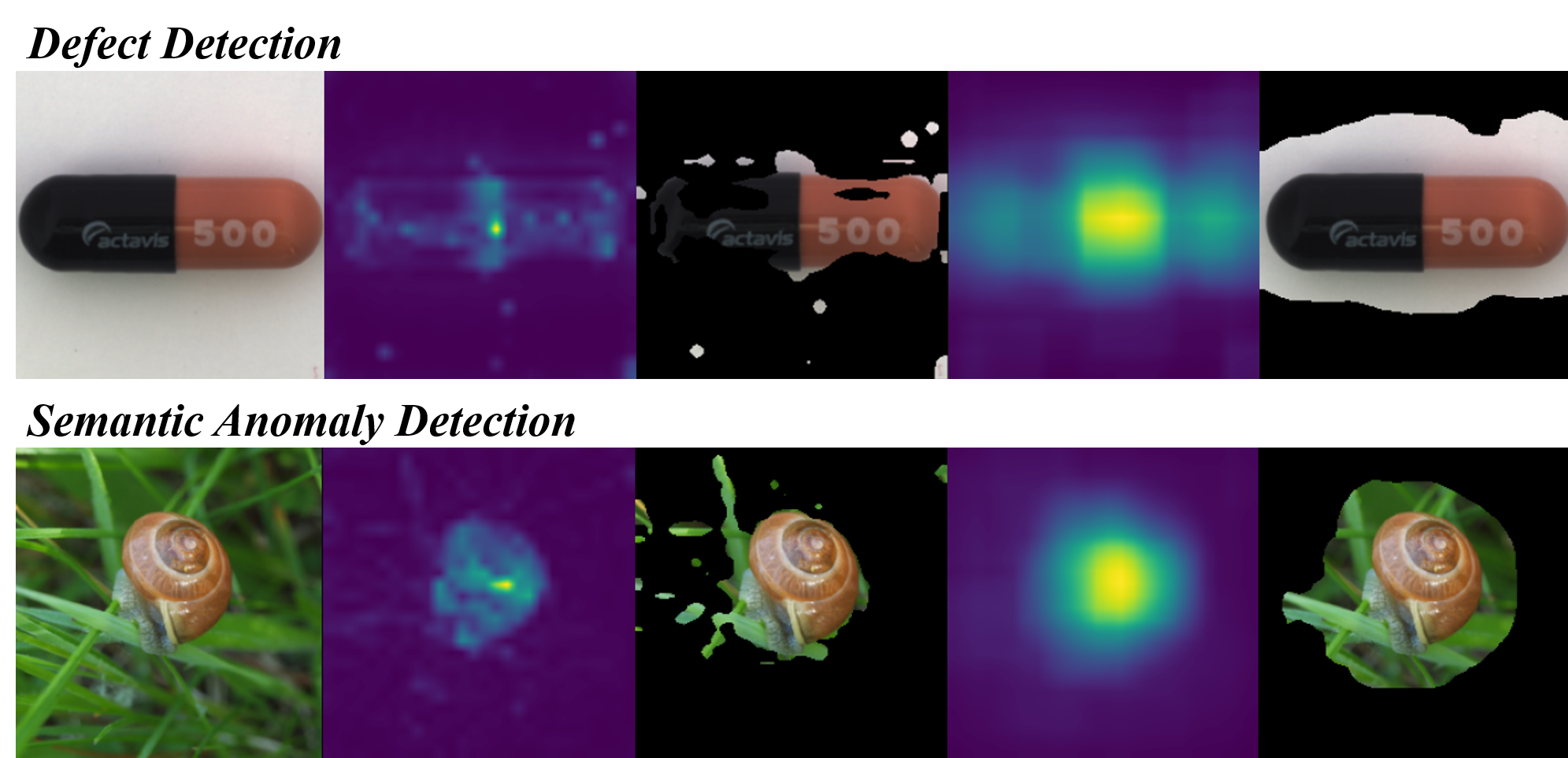}
    \end{center}
        \caption{\textbf{Back Patch Masking with smoothed attention.} The sequence from left to right shows the original image, raw-attention, raw-attention masked image, smoothed attention, and smoothed-attention masked image with boundaries set by a threshold. The threshold is set to 0.1.}
        \label{fig:backpatch_masking}
\end{figure}

To overcome these challenges, alternative target-aware detection approaches need to be explored. Drawing inspiration from the inherent object boundaries in the self-attention map of self-supervised Vision Transformers (ViTs) (Fig.~\ref{fig:backpatch_masking}), we leverage this characteristic to develop a target-aware detection framework. We generate a pseudo mask $\mathrm{M}$ from the attention map of self-supervised ViTs to segment the foreground, including the region of interest (RoI), and apply it to a set of patches from the $i$-th image, $\mathcal{P}_i$. Specifically, we calculate the self-attention of the $\texttt{[CLS]}$ token in the last layer of $L$ and average it across heads, resulting in an attention map $\mathrm{S}_A \in \mathbb{R}^{\frac{H}{p}\times\frac{W}{p}}$, where $H$ and $W$ denote the height and width of an input image, respectively. 
   
   However, the attention map tends to be rather coarse, often encompassing non-RoI areas, and may fail to effectively filter out background noise, as depicted in Fig.~\ref{fig:backpatch_masking}. To address this concern, we introduce attention smoothing to reduce the risk of false positive cases. For attention smoothing, we utilize a smoothing kernel $g(n) = n^{-2}J_n$ with a kernel size of $n$. Subsequently, we apply $g(n)$ to $\mathrm{S}_A^{(i,j)}$ at position $(i,j)$ according to the following equation:
   
   \begin{equation}
   \mathrm{S}_A^{(i,j)} = \sum_{a}\sum_{b}{\mathrm{S}_A^{(i+a,j+b)}g(n)}, \quad a,b \in (-n, n)
   \label{eq:attention_smoothing}
   \end{equation}
   
   The resulting smoothed attention map serves as the aforementioned pseudo mask $\mathrm{M} \in \mathbb{R}^{\frac{H}{p} \times \frac{W}{p}}$ (Fig.~\ref{fig:backpatch_masking}). While target-aware detection guidance through attention proves valuable in preventing false positives, it may introduce vulnerability to false negatives when no instances are present. Hence, instead of employing soft masking as commonly found in the literature, we binarize $\mathrm{M}$ using a threshold to remove irrelevant regions. Finally, we apply this to $\mathcal{P}_i$ as $\mathcal{P}_{mask, i} = \mathrm{M} \otimes \mathcal{P}_i \in \mathbb{R}^N$. Back Patch Masking (BPM) provides effective target-aware guidance by ensuring that regions below a certain threshold are not mistakenly rejected while attending to target areas. Furthermore, in the training stage, BPM drops the patches from the non-target area and reduces the size of the patch-level memory bank $\mathcal{M}$. This allows us to achieve unified and robust anomaly detection, bridging the gap between defect detection and semantic anomaly detection (especially in the multi-class case).


\subsection{Top \textbf{\textit{k}}-ratio Feature Matching}\label{sec:tkfm}
 We discover that task unification can be achieved by adopting the patch-level anomaly learning approach and incorporating multiple instance learning (MIL) into anomaly scoring. In patch-level anomaly learning, it is a common practice to aggregate patch-level score maps to compute an image-level score map. Two aggregation approaches are commonly used: the max operation, which determines the anomaly based on the patch with the highest anomaly score \cite{roth2022towards}, and top-$k$ MIL aggregations \cite{ding2022catching}. The former has the drawback of reduced robustness as it relies on a single element to make judgments for the entire image. In contrast, the latter, which utilizes top-$k$ MIL, addresses this vulnerability.
    
Our approach shares similarities with the top-$k$ MIL approach. In a previous study \cite{ding2022catching}, a supervised setting is employed where the features of the top-$k$ anomalous image patches are updated based on their corresponding labels (anomalous or not). In contrast, we solely rely on the anomaly score without undergoing any training process. Additionally, the focus of \cite{ding2022catching} is specifically on defect detection, whereas our approach extends the top-$k$ MIL approach to integrate the handling of various levels of anomaly detection, from local to semantic.
    
To compute the image-level anomaly score, we first compute the patch-wise distance $\mathcal{D}_{N}$ between a test image $d^{test}$ and normal patches in $\mathcal{M}$. Then, we reorder $\mathcal{D}_{N}$ in descending order (i.e., order by distance from normal patches) and set $\mathcal{S}_{(i)} = \mathcal{D}_{(N-i+1)}$. We consider only the top $k$\% of reverse-ordered patch distances $\mathcal{S}_{(i)}$ for image-level anomaly score as $\mathcal{S}_{(K)} = \{s_{(1)}, s_{(2)}, ..., s_{(K)}\}$. Finally, an image-level anomaly score $s^*$ is as follows:

\begin{equation}
s^{*} = \frac{1}{K}\sum_{i=1}^{N}{\mathbbm{1}(i \le K)}\mathcal{S}_{mask,(i)},
\label{eq:score}
\end{equation}
    
where $K$ = $\lceil{(k/100)*N}\rceil$ and $\mathcal{S}_{mask}$ denotes a score map from masked patches $\mathcal{P}^{test}_{mask,i}$. By conducting anomaly scoring as an ensemble of multiple instances, we can achieve unified anomaly detection in a task-agnostic manner across various tasks and datasets. Furthermore, our experimental validation has confirmed that this top $k$-ratio feature ensemble can be extended beyond binary detection to perform grouping of slight anomalies (Sec. \ref{sec:topk_validation}).

\section{Experiments}\label{sec:experiments}
To demonstrate that $UniFormaly$ is a powerful and general anomaly detection framework, We evaluate \textit{UniFormaly} in various anomaly detection scenarios across anomaly levels and task objectives - Defect detection and localization, semantic anomaly detection, anomaly clustering (Sec.\ref{sec:main_results}), and low shot anomaly detection(Sec.\ref{sec:low_shot}). In addition, we report results on both one-class and multi-class normality scenarios, showing that ours work well in both settings. Lastly, we empirically prove that our components - BPM and Top $k$-ratio feature matching play a key role in \textit{UniFormaly} (Sec.\ref{sec:ablation}). Regarding reported results, we mark the best results in bold and underline the second-best results unless otherwise specified.

\subsection{Experimental Details}\label{sec:experimental_details}

\noindent{\bf Datasets.} Our defect detection experiments primarily utilize the widely-used benchmark MVTecAD~\cite{bergmann2019mvtec}, which contains 15 sub-datasets with different categories, including 10 objects and 5 textures. In addition, we conduct experiments on Magnetic Tile Defects (MTD)~\cite{huang2020surface}, BeanTech Anomaly Detection (BTAD)~\cite{mishra2021vt}, and Casting manufacturing Product Defect (CPD) to demonstrate the scalability of \textit{UniFormaly} to more specialized tasks. For semantic anomaly detection, we perform experiments on CIFAR-10/100~\cite{krizhevsky2009learning}, ImageNet-30~\cite{hendrycks2019using}, and Species-60, the sub-datasets of Species~\cite{hendrycks2022scaling}. Species is a large-scale and high-resolution dataset of which labels are non-overlapped with ImageNet, comprising over 700,000 images of a thousand subcategories. We randomly select 60 subcategories with more than 500 images and prepare the train-test split following ~\cite{hendrycks2022scaling}.

\noindent{\bf Implementation details.} We obtained patch embeddings from layers $L = \{3,...,10\}$ in two self-supervised ViT architectures: DINO ViT-B/8 \cite{caron2021emerging} and iBOT ViT-B/16 \cite{zhou2021ibot}. These configurations are denoted as \textit{UniFormaly}-DINO and \textit{UniFormaly}-iBOT, respectively. To smooth the attention map in our Back Patch Masking, we utilize the attention scores from the last layer and apply a smoothing kernel with a size of $n=7$. We maintain a consistent threshold of 0.1 for binary masking across all tasks, as empirical findings have shown its effectiveness across various datasets and tasks. Unless specified otherwise, we set the top $k$-ratio to 5\% for all tasks. These uniform settings ensure our system operates in a task-agnostic manner. Additionally, we refrain from fine-tuning all methods on the data, adhering to unsupervised settings that assume a lack of abnormal data during training.
    
\noindent{\bf Evaluation metrics.} We use the threshold-independent metric AUROC to evaluate image-level and pixel-level anomaly detection performance. For anomaly clustering, we employ NMI (Normalized Mutual Information) and ARI (Adjusted Random Index), two popular metrics for clustering quality analysis when ground-truth cluster assignments are available for the test set. Furthermore, to address label imbalance, we report the F1 score after matching ground truth and cluster prediction using the Hungarian method~\cite{kuhn1955hungarian}.

\subsection{Main Results}\label{sec:main_results}

\noindent{\bf Defect detection.} In Table \ref{tab:mvtec_res}, we present a comprehensive evaluation result for anomaly detection and localization across various approaches, including unsupervised~\cite{zavrtanik2021draem,wu2021learning}, self-supervised~\cite{yi2020patch,li2021cutpaste}, and pre-trained~\cite{defard2021padim,roth2022towards} on the MVTecAD. The results demonstrate that \textit{UniFormaly} outperforms or shows competitive results compared to its baselines. Additionally, our method shows significantly less standard deviation across classes than most of the baselines, indicating our applicability to various sub-datasets with distinct categories. Furthermore, Table~\ref{tab:defect_other_res} presents the results on other benchmarks (MTD, BTAD, CPD), demonstrating that our method outperforms the state-of-the-art methods in all datasets. This confirms that, compared to pre-trained encoder-based methods, the methods with proxy tasks exhibit larger performance fluctuations with changes in datasets. In contrast, \textit{UniFormaly} consistently presents outstanding performance without fine-tuning, demonstrating its robustness in various applications.

\begin{table*}
\begin{center}
\caption{\textbf{Defect detection on MVTecAD.} We present the image-level AUROC and pixel-level AUROC. Pixel-level AUROC for MetaFormer is not reported as it is unavailable in the original paper. We report the mean and standard deviation of 15 classes in MVTecAD.}
\resizebox{1.0\textwidth}{!}{
\begin{tabular}{@{}cc|cccccc|cc@{}}
\toprule[1.0pt]
\multicolumn{2}{c|}{\multirow{2}{*}{}}                                                             & \multirow{2}{*}{PSVDD~\cite{yi2020patch}} & \multirow{2}{*}{CutPaste~\cite{li2021cutpaste}} & \multirow{2}{*}{PaDiM~\cite{defard2021padim}} & \multirow{2}{*}{MetaFormer~\cite{wu2021learning}} & \multirow{2}{*}{DRAEM~\cite{zavrtanik2021draem}} & \multirow{2}{*}{PatchCore~\cite{roth2022towards}} & \multicolumn{2}{c}{Ours} \\ \cmidrule(l){9-10} 
\multicolumn{2}{c|}{}                                                                              &                        &                           &                        &                             &                        &                            & iBOT        & DINO       \\ \midrule\midrule
\multicolumn{1}{c|}{\multirow{2}{*}{\begin{tabular}[c]{@{}c@{}}Image\\ AUROC\end{tabular}}} & Mean & 92.07                   & 95.16                      & 95.42                   & 95.81                        & 97.98                   & 99.07                       & \underline{99.11}        & \textbf{99.32}       \\
\multicolumn{1}{c|}{}                                                                       & Std  & 6.30                    & 5.02                       & 4.15                    & 4.41                         & 2.67                    & \textbf{0.98}                        & 1.07         & \underline{1.02}        \\ \midrule
\multicolumn{1}{c|}{\multirow{2}{*}{\begin{tabular}[c]{@{}c@{}}Pixel\\ AUROC\end{tabular}}} & Mean & 95.71                   & 95.71                     & 97.41                   & -                           & 97.31                   & 98.07                       & \underline{98.30}        & \textbf{98.48}       \\
\multicolumn{1}{c|}{}                                                                       & Std  & 2.29                    & 2.90                       & 1.57                    & -                           & 2.43                    & 1.32                        & \textbf{0.82}         & \underline{0.95}        \\ \bottomrule[1.0pt]
\end{tabular}
\label{tab:mvtec_res}
}
\end{center}
\end{table*} 

\begin{table}
\caption{\textbf{Defect detection on other benchmarks.} Our baselines are state-of-the-art frameworks from unsupervised~\cite{zavrtanik2021draem}, self-supervised~\cite{li2021cutpaste}, and pre-trained~\cite{roth2022towards} methods. We obtained the baseline results for other benchmarks by applying identical settings to those used in MVTecAD and using their released code. However, for CutPaste, we produced the results ourselves since the official code was not available.}
    \begin{center}
    \resizebox{\linewidth}{!}{
    \begin{tabular}{@{}c|ccc|cc@{}}
    \toprule[1.0pt]
    \multirow{2}{*}{} & \multirow{2}{*}{DRAEM~\cite{zavrtanik2021draem}} & \multirow{2}{*}{CutPaste~\cite{li2021cutpaste}} & \multirow{2}{*}{PatchCore~\cite{roth2022towards}} & \multicolumn{2}{c}{Ours}      \\ \cmidrule(l){5-6} 
                      &                        &                           &                            & iBOT          & DINO          \\ \midrule\midrule
    MTD               & 70.3                   & 93.8                      & 97.6                       & \textbf{99.1} & \underline{99.0} \\
    BTAD              & 85.3                   & 93.1                      & 92.9                       & \textbf{94.9} & \underline{94.8} \\
    CPD               & 92.8                   & 97.4                      & 96.8                       & \underline{99.3} & \textbf{99.8} \\ \bottomrule[1.0pt]
    \end{tabular}
    }
    \end{center}
    \label{tab:defect_other_res}
    \end{table}

    \noindent{\bf Semantic anomaly detection.} In Table~\ref{tab:semantic_res}, we observe that our method consistently outperforms baselines. Particularly in high-resolution datasets such as ImageNet-30 and Species-60, our method exhibits a significant performance margin compared to the alternatives. In low-resolution datasets, such as CIFAR-10 and CIFAR-100, our method shows less improvement in AUROC compared to the other datasets since ViTs are pre-trained on the resolution of 224 (which is much higher than that of CIFAR-10/100). While a resolution discrepancy exists between the test images and the pretraining datasets, UniFormaly-DINO shows competitive results, and UniFormaly-iBOT even outperforms the previous state-of-the-art (SOTA) without requiring any adaptation to the test images. We acknowledge potential fairness concerns when comparing with ImageNet-30 since \textit{UniFormaly} is built on an ImageNet-pre-trained encoder with self-supervision. To address this, we conduct experiments on Species-60, a dataset without semantic overlap with ImageNet. As shown in Table~\ref{tab:semantic_res}, for ImageNet-30 and Species-60, our method outperforms its competitors by a large gap. This confirms the effectiveness of our approach in detecting semantic anomalies and slight anomalies, such as defects, regardless of the aforementioned overlap issue. Meanwhile, both~\cite{hendrycks2019using} and~\cite{tack2020csi}, which are self-supervised approaches trained from scratch on the given dataset, exhibit a more significant difference in AUROC compared to ours. This result suggests that the training setups of baselines are tailored to the ImageNet-30 and suboptimal for Species-60. Moreover, PatchCore, another pre-trained encoder-based method, performs poorly on semantic anomaly detection due to an architectural design tailored to defect detection.

    \begin{table}
    \caption{\textbf{Semantic anomaly detection.} PatchCore's AUROC for ImageNet-30 is not reported because it relies on a supervised pre-trained encoder with label information from ImageNet. We produced the results of our competitors for Species-60 using the same setting as for ImageNet-30.}
    \begin{center}
    \resizebox{\linewidth}{!}{
    \begin{tabular}{@{}c|ccc|cc@{}}
    \toprule[1.0pt]
    \multirow{2}{*}{} & \multirow{2}{*}{SS-OOD~\cite{hendrycks2019using}} & \multirow{2}{*}{CSI~\cite{tack2020csi}} & \multirow{2}{*}{PatchCore~\cite{roth2022towards}} & \multicolumn{2}{c}{Ours} \\ \cmidrule(l){5-6} 
                      &                         &                      &                            & iBOT       & DINO        \\ \midrule\midrule
    CIFAR-10          & 89.8                    & \underline{94.3}                 & 88.5                          & \textbf{94.7}       & 92.9       \\
    CIFAR-100         & 79.8                    & 89.6                 & 82.9                          & \textbf{91.2}       & \underline{90.0}           \\
    ImageNet-30       & 85.7                    & 91.6                 & -                          & \textbf{97.6}       & \underline{96.8}        \\
    Species-60        & 74.8                    & 81.4                 & 82.4                       & \textbf{94.4}       & \underline{92.6}        \\ \bottomrule[1.0pt]
    \end{tabular} 
    }
    \end{center}
    \label{tab:semantic_res}
    \end{table}

\begin{table*}
\begin{center}
\caption{\textbf{Results for Multi-class defect detection on the MVTecAD Datasets.} One-class detection results are given in parentheses.}
\resizebox{\textwidth}{!}{
\begin{tabular}{@{}c|ccccccc|cc@{}}
\toprule[1.5pt]
\multirow{2}{*}{} & \multirow{2}{*}{US\cite{bergmann2020uninformed}} & \multirow{2}{*}{PSVDD\cite{yi2020patch}} & \multirow{2}{*}{CutPaste\cite{li2021cutpaste}} & \multirow{2}{*}{PaDiM\cite{defard2021padim}} & \multirow{2}{*}{DRAEM\cite{zavrtanik2021draem}}  & \multirow{2}{*}{Patchcore\cite{roth2022towards}}  & \multirow{2}{*}{UniAD\cite{lu2022unified}} & \multicolumn{2}{c}{Ours}    \\ \cmidrule(l){9-10} 
                  &                     &                        &                       &                           &                        &                      &                        &                         iBOT         & DINO         \\ \midrule\midrule
Image             & 74.5(87.7)          & 76.8(92.1)                                 & 77.5(96.1)                & 84.2(95.5)                  & 88.1(98.0)    & \underline{98.8(99.1)}         & 96.5(96.6)             & \underline{98.8 (99.1)} & \textbf{99.2(99.3)} \\
Pixel             & 81.8(93.9)          & 85.6(95.7)                      & -                         & 89.5(97.4)                   & 87.2(97.3)   &97.8(98.1)          & 96.8(96.6)             & \underline{98.0 (98.3)}     & \textbf{98.1(98.5)} \\ \bottomrule[1.5pt]
\end{tabular}
\label{tab:multiclass_res}
}
\end{center}
\end{table*}

\begin{table}[ht!]
\begin{center}
\caption{\textbf{Multi-class Semantic anomaly detection results on CIFAR-10 Datasets.}}
\centering
\resizebox{1.0\linewidth}{!}{%
\begin{tabular}{@{}c|ccccc|cc@{}}
\toprule[1.5pt]
\multirow{2}{*}{\thead{Normal\\Indices}} & \multirow{2}{*}{US~\cite{bergmann2020uninformed}} & \multirow{2}{*}{FCDD~\cite{liznerski2021explainable}} & \multirow{2}{*}{FCDD+OE~\cite{liznerski2021explainable}} & \multirow{2}{*}{PANDA~\cite{reiss2021panda}} & \multirow{2}{*}{UniAD~\cite{lu2022unified}} & \multicolumn{2}{c}{Ours} \\ \cmidrule(l){7-8} 
                                &                     &                       &                          &                                         &                        & iBOT        & DINO       \\ \midrule\midrule
\{01234\}                       & 51.3                & 55.0                  & 71.8                     & 66.6                                 & \textbf{84.4}                   & \underline{79.9}        & 72.8       \\
\{56789\}                       & 51.3                & 50.3                  & 73.7                     & 73.2                                  & \underline{80.9}                   & \textbf{84.7}        & 80.2       \\
\{02468\}                       & 63.9                & 59.2                  & 85.3                     & 77.1                               & \textbf{93.0}                   & \underline{90.7}        & 85.8       \\
\{13579\}                       & 56.8                & 58.5                  & 85.0                     & 72.9                              & \textbf{90.6}                   & \underline{86.4}        & 83.2          \\ \midrule
Mean                            & 55.9                & 55.8                  & 78.9                     & 72.4                                & \textbf{87.2}                   & \underline{85.4}        & 80.5          \\ \bottomrule[1.5pt]
\end{tabular}
}
\label{tab:multiclass_cifar}
\end{center}
\end{table}

\begin{table}[ht!]
\begin{center}
\caption{\textbf{Results for Multi-class Semantic anomaly detection on the Species-60 Datasets.} We conducted tests for all methods using 5 random splits, with each split comprising 30 normal classes. Additional split details can be found in the supplementary material.}
\scriptsize{
\resizebox{\linewidth}{!}{
\begin{tabular}{@{}c|ccc|cc@{}}
\toprule[1.0pt]
\multirow{2}{*}{\thead{Normal\\ Split}} & \multirow{2}{*}{UniAD~\cite{lu2022unified}} & \multirow{2}{*}{PANDA~\cite{reiss2021panda}} & \multirow{2}{*}{PatchCore~\cite{roth2022towards}} & \multicolumn{2}{c}{Ours}     \\ \cmidrule(l){5-6} 
                  &                        &                        &                            & iBOT                  & DINO \\ \midrule\midrule
A           & 55.8                   & 57.6                   & 73.9                       & \textbf{82.6}                  & \underline{78.5}    \\
B           & 53.4                      & 58.6                      & 74.1                          & \textbf{79.6}                  & \underline{76.1}    \\
C           & 55.2                      & 53.4                      & 72.3                          & \textbf{80.2}                  & \underline{76.5}    \\
D           & 50.3                      & 56.4                      & \underline{74.4}                          & \textbf{79.6}                  & 74.3    \\
E           & 50.0                      & 54.0                      & 70.2                          & \textbf{78.1}                  & \underline{72.4}    \\ \midrule
Mean              & 52.9                      & 56.0                      & 73.0                          & \textbf{80.0} & \underline{75.6}    \\ \bottomrule[1.0pt]
\end{tabular}
}
\label{tab:multiclass_species}
}
\end{center}
\end{table}
\noindent{\bf Multi-class anomaly detection.} Our method is easily scalable to multi-class anomaly detection by incorporating representations of multi-class data in a normal memory bank. We present multi-class defect detection results on MVTecAD in Table \ref{tab:multiclass_res}, as well as semantic detection results on CIFAR-10 and Species-60 in Table \ref{tab:multiclass_cifar} and \ref{tab:multiclass_species}, respectively. In Table~\ref{tab:multiclass_res}, we observe that in multi-class defect detection, our approach shows almost no performance drop in AUROC. The method proposed for the multi-class detection task~\cite{you2022a} also exhibits a minimal performance drop compared to the other methods, except for ours. UniAD~\cite{you2022a} consistently lags behind our performance in defect detection overall and shows poor performance on Species-60, failing to converge. These observations suggest that the learning-based multi-class detection framework~\cite{you2022a} may not be successfully applicable to semantic anomaly detection with more complicated and high-resolution images, indicating a lack of generality. Similar convergence issues are observed in \cite{reiss2021panda}. On the contrary, pre-trained encoder-based methods, including PatchCore\cite{roth2022towards} and ours, alleviate these issues, as shown in Table \ref{tab:multiclass_species}. In addition, \textit{UniFormaly}-DINO achieves a new state-of-the-art performance in both one-class and multi-class settings for defect detection (Table \ref{tab:multiclass_res}). Our method also demonstrates the same result for anomaly localization, achieving state-of-the-art results with only a minimal performance drop of 0.1 with \textit{UniFormaly}-DINO. In multiclass semantic anomaly detection, our method achieves the highest score on Species-60 (Table \ref{tab:multiclass_species}), while still delivering competitive results on CIFAR-10 (Table \ref{tab:multiclass_cifar}). Although the previous pre-trained encoder-based method\cite{roth2022towards} shows relatively high scores compared to the other online learning methods, it still falls behind us by 7 in AUROC.

\noindent{\bf Anomaly clustering.} \textit{UniFormaly} effectively clusters anomalies into categories using the top $k$\% of patch embeddings based on reordered features, capturing more than just binary information of normality. We easily apply our averaged features to K-means clustering. We compare our method with~\cite{sohn2023anomaly}, which is the only method that has studied anomaly clustering task, as well as the deep clustering methods, IIC~\cite{ji2019invariant}, GATCluster~\cite{niu2020gatcluster}, and SCAN~\cite{van2020scan} as baselines. Although our approach is not primarily designed for clustering, it outperforms baselines, including \cite{sohn2023anomaly} (Table \ref{tab:clustering}). In terms of F1 scores, we improve the strongest baseline by 10.48\% and achieve the highest F1 score. In particular, with K-means clustering, we outperform the state-of-the-art method~\cite{sohn2023anomaly}.


\begin{table}
\caption{\textbf{Clustering results on MVTecAD.} We obtained the baseline method results from \cite{sohn2023anomaly}. For AC\cite{sohn2023anomaly}, to ensure a fair comparison, we report the results of K-Means Clustering.}
\centering
\resizebox{\linewidth}{!}{
    \begin{tabular}{@{}c|cccc|lc@{}}
    \toprule[1.0pt]
    \multirow{2}{*}{} & \multirow{2}{*}{IIC~\cite{ji2019invariant}} & \multirow{2}{*}{GATCluster~\cite{niu2020gatcluster}} & \multirow{2}{*}{SCAN~\cite{van2020scan}} & \multirow{2}{*}{AC~\cite{sohn2023anomaly}} & \multicolumn{2}{c}{Ours}                  \\ \cmidrule(l){6-7} 
                      &                      &                             &                       &                     & \multicolumn{1}{c}{iBOT} & DINO           \\ \midrule\midrule
    NMI               & 0.093                & 0.136                       & 0.210                 & 0.500               & \underline{0.528}           & \textbf{0.547} \\
    ARI               & 0.020                & 0.053                       & 0.103                 & 0.390               & \underline{0.423}           & \textbf{0.433} \\
    F1                & 0.285                & 0.264                       & 0.335                 & 0.601               & \underline{0.643}           & \textbf{0.645} \\ \bottomrule[1.0pt]
    \end{tabular}
    }
\label{tab:clustering}
\end{table}

\subsection{Low-shot Anomaly Detection}\label{sec:low_shot}

\begin{figure}[ht!]
    \begin{center}
        \includegraphics[width=1.0\linewidth]{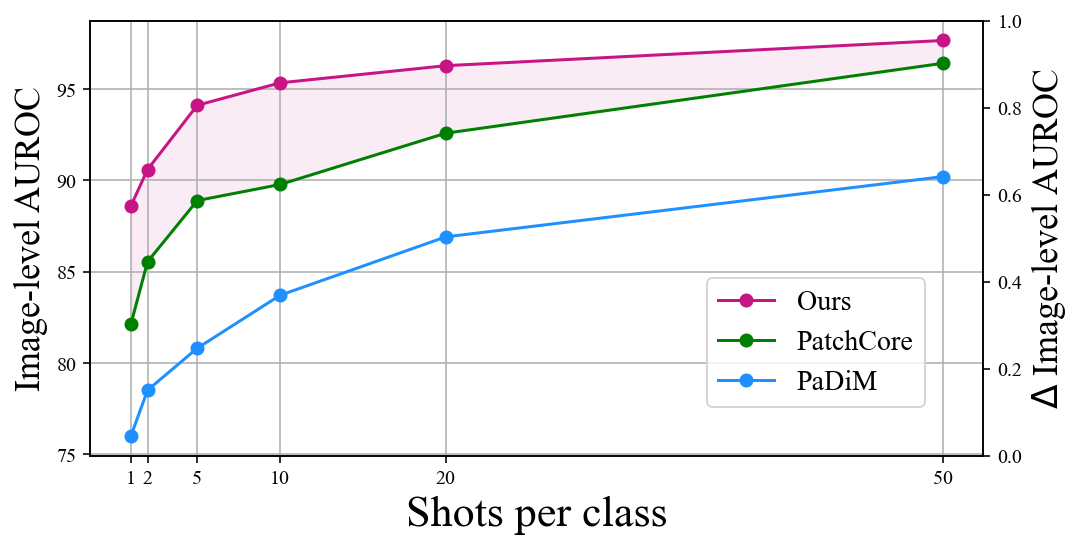}
    \end{center}
        \caption{\textbf{Low-shot anomaly detection result.} We report our results as the average of \textit{UniFormaly}-DINO and \textit{UniFormaly}-iBOT.}
    \label{fig:lowshot}
    \end{figure}

    To assess the performance of our unified system in scenarios with limited normal data, we conducted low-shot anomaly detection experiments. Even in scenarios with only a few normal shots, \textit{UniFormaly} outperforms other pre-trained methods, as depicted in Fig.~\ref{fig:lowshot}. Despite previous pre-trained encoder-based methods performing well in the low-shot setting, our approach achieves an AUROC of approximately 90 in the one-shot setting, improving the previous state-of-the-art by approximately 8\%. Even with only 50 images, a quarter of the normal training data, \textit{UniFormaly} achieves a competitive AUROC of 98, matching the previous state-of-the-art performance (see Fig.~\ref{fig:lowshot}). This demonstrates the cost-effectiveness of our method for building datasets and its effectiveness in cases where only a limited number of sample images are available.

\subsection{Ablation Study}\label{sec:ablation}
    We conduct ablation studies to verify the critical role of BPM and Top $k$-ratio feature matching in developing a unified anomaly detection system. The results are reported with \textit{UniFormaly}-DINO on the MVTecAD for defect detection, multi-class anomaly detection, and anomaly clustering, while for semantic anomaly detection, results are reported on Species-60.

    \subsubsection{Top $k$-ratio feature matching}\label{sec:topk_validation}
    
\begin{figure}[ht!]
    \begin{center}
    \includegraphics[width=\linewidth]{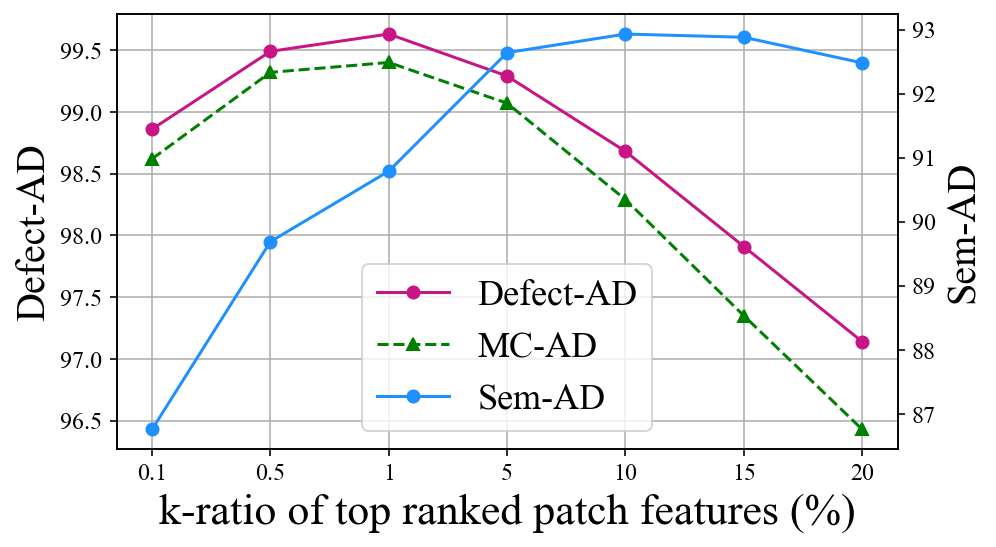}
    \end{center}
    \caption{\textbf{Results on top k(\%)-ratio with varying k.} We evaluate image-level AUROC varying $k$ on defect detection (Defect-AD), multi-class anomaly detection (MC-AD), and semantic anomaly detection (Sem-AD). }
    \label{fig:topk_scores}
    \end{figure}

    In Fig.~\ref{fig:topk_scores}, we investigate the effect of $k$-ratio on various tasks by evaluating changes in image-level AUROC over varying $k$ from 0.1 to 20. The results confirm that the top $k$-ratio scores enable the detection of anomalies of various levels by referring to the multiple high-scored regions. When $k$ is set to 0.1, representing the case without top $k$-ratio feature matching applied, a substantial performance gap exists between semantic anomaly detection and other tasks, indicating a lack of task generality. However, when top $k$-ratio feature matching is applied ($k$ is larger than 0.1), individual task performances improve, and the performance gap between tasks diminishes. Regardless of the different shapes of the AUROC curve for $k$ due to varying anomaly levels for each task, all tasks achieve outstanding performance at around 5\%, highlighting the effectiveness of top $k$-ratio feature matching for task-agnostic anomaly detection.

    We further compare three cases—All, Max, and Top $k$-ratio—regarding the selection of features in anomaly clustering, and the results demonstrate the superiority of top $k$-ratio feature matching. Table \ref{tab:topk_clustering} demonstrates that top $k$-ratio outperforms other scenarios for both $k$=1 and 5. Specifically, when utilizing 5\% of patch features, ARI shows improvements of 57.2\% and 14.4\% over `All' and `Max,' respectively. Even with $k$=1, the margin widens further, resulting in ARI improvements of 83\% and 33.2\% over `All' and `Max,' respectively. Anomaly clustering for defect types requires capturing fine-grained details in semantically similar but locally distinct images, making global representations unsuitable. The Max case, relying on a single patch with the highest anomaly score, is less effective due to its vulnerability to false positives. In contrast, the top $k$-ratio approach utilizes patch-level representations, capturing fine-grained details while remaining robust against false positives, resulting in improved clustering performance.

    \begin{table}
    \begin{center}
    \caption{\textbf{Effect of top $k$-ratio feature matching in anomaly clustering on MVTecAD.} In each scenario, we apply clustering algorithms to three types of features: the average-pooled feature of all patch-level representations, the feature with the highest anomaly score, and the average-pooled feature of patch features with the top $k$\% of anomaly scores.}
    \scriptsize{
    \renewcommand{\arraystretch}{0.65}
        \resizebox{\linewidth}{!}{
            \begin{tabular}{@{}c|c|c|cc@{}}
            \toprule
            \multirow{2}{*}{} & \multirow{2}{*}{All} & \multirow{2}{*}{Max} & \multicolumn{2}{c}{\textbf{Top $k$-ratio}}            \\ \cmidrule(l){4-5} 
                              &                      &                      & \multicolumn{1}{c|}{$k$=5} & {$k$=1}    \\ \midrule
            \multicolumn{1}{c||}{NMI}               & 0.3916               & 0.5204               & \multicolumn{1}{c|}{{\underline {0.547}}} & \textbf{0.6127} \\
            \multicolumn{1}{c||}{ARI}               & 0.2757               & 0.3790               & \multicolumn{1}{c|}{{\underline {0.4334}}} & \textbf{0.5048} \\
            \multicolumn{1}{c||}{F1}                & 0.5236               & 0.6042               & \multicolumn{1}{c|}{{\underline {0.6447}}} & \textbf{0.7072} \\ \bottomrule
            \end{tabular}
        }
    }
    \label{tab:topk_clustering}
    \end{center}
    \end{table}

\subsubsection{Back Patch Masking}

\noindent{\bf Smoothing kernel size $n$.} Table \ref{tab:bpm_smoothing} demonstrates the impact of smoothing kernel size $n$ on Back Patch Masking. When using a raw attention map with $n$ set to 1, inferior results are obtained. This is significant in the case of defect detection, semantic anomaly detection, and multiclass anomaly detection. This finding aligns with the previous discussion, where we noted that a raw attention map is highly susceptible to anomalous-like normal regions. Based on our results, we observe that consistent performance is achieved regardless of the chosen kernel size $n$. As a result, we have chosen a default kernel size $n$ of 7, which serves as a reasonable compromise and works effectively across various tasks.

 \begin{table}[ht!]
\begin{center}
\caption{\textbf{Results on varying smoothing kernel size $n$.} We evaluate the performance of our method on multiple tasks with varying $n$ from 1 to 13 by 2. We marked the results at our default settings in bold.}
\resizebox{\linewidth}{!}{
\begin{tabular}{@{}cccccccc@{}}
\toprule[1.0pt]
    $n$  & 1      & 3      & 5      & 7      & 9      & 11     & 13     \\ \midrule
\multicolumn{8}{l}{\textit{\textbf{Defect Detection}}}               \\
\rowcolor[HTML]{FFD5D5} 
Image & 93.00  & 98.15   & 99.31  & \textbf{99.32}  & 99.25  & 99.33  & 99.35  \\
\rowcolor[HTML]{FFD5D5} 
Pixel & 94.12  & 97.4  & 98.26  & \textbf{98.48}  & 98.48  & 98.51  & 98.51  \\ \midrule
\multicolumn{8}{l}{\textit{\textbf{Semantic Anomaly Detection}}}     \\
\rowcolor[HTML]{DEF5F8} 
Image & 88.63  & 92.14   & 92.54  & \textbf{92.64}  & 92.51  & 92.49  & 92.42   \\ \midrule
\multicolumn{8}{l}{\textit{\textbf{Multiclass Anomaly Detection}}}   \\
\rowcolor[HTML]{FFF2CC} 
Image & 91.33  & 97.55  & 99.19  & \textbf{99.15}  & 99.20   & 99.20  & 99.19  \\
\rowcolor[HTML]{FFF2CC} 
Pixel & 93.84  & 97.23  & 98.08  & \textbf{98.33}  & 98.34  & 98.29  & 98.19  \\ \midrule
\multicolumn{8}{l}{\textit{\textbf{Anomaly Clustering}}}             \\
\rowcolor[HTML]{D9D2E9} 
NMI   & 0.5038 & 0.538 & 0.5494 & \textbf{0.547} & 0.5416 & 0.5390 & 0.5446 \\
\rowcolor[HTML]{D9D2E9} 
ARI   & 0.3994 & 0.4355 & 0.4522 & \textbf{0.4334} & 0.4338 & 0.4291 & 0.4362 \\
\rowcolor[HTML]{D9D2E9} 
F1    & 0.6229 & 0.6405 & 0.6687 & \textbf{0.6447} & 0.6422 & 0.6377 & 0.6387 \\ \bottomrule[1.0pt]
\end{tabular}
}
\label{tab:bpm_smoothing}
\end{center}
\end{table}

\noindent{\bf Binary masking threshold $\lambda$.} To examine the impact of binary masking with a threshold on Back Patch Masking, we compare it with soft masking, where the original representations are multiplied by the floating-point values of the raw attention scores ranging from 0 to 1. Table~\ref{tab:bpm_thres} shows that soft masking is significantly less effective than binary masking in most cases. This highlights the critical role played by binary masking in Back Patch Masking. Moreover, we investigate the effect of varying threshold value $\lambda$ on Back Patch Masking performance. Our results demonstrate that a threshold value of $\lambda=0.1$ performs best across most of the tasks considered, with a slight decrease in performance observed as $\lambda$ increases beyond this value. Nevertheless, \textit{UniFormaly} consistently performs well across various tasks, even with varying binary masking thresholds.

\begin{table}[ht!]
\begin{center}
\caption{\textbf{Effect of binary masking and its threshold $\lambda$.} We show the efficacy of binary masking compared to soft masking. To further investigate the effect of binary masking threshold $\lambda$, we present the results for our tasks with varying $\lambda$ from 0.1 to 0.3. We marked the results at our default settings in bold.}
\scriptsize{
    \resizebox{\linewidth}{!}{
    \begin{tabular}{@{}cc|ccc|c@{}}
    \toprule[1.0pt]
    & & \multicolumn{3}{c|}{Binary} & \multirow{2}{*}{Soft} \\ \cmidrule(lr){3-5}
    & & 0.1 & 0.2 & 0.3 & \\ \midrule
    \rowcolor[HTML]{FFD5D5} 
    \cellcolor[HTML]{FFD5D5} & Image & \textbf{99.32} & 99.10 & 98.84 & 97.38 \\
    \rowcolor[HTML]{FFD5D5} 
    \multirow{-2}{*}{\cellcolor[HTML]{FFD5D5}\textit{Defect-AD}} & Pixel & \textbf{98.47} & 97.08 & 94.49 & 94.88  \\ \midrule
    \rowcolor[HTML]{DEF5F8} 
    \textit{Sem-AD} & Image & \textbf{92.64} & 93.04 & 93.38 & 90.49  \\ \midrule
    \rowcolor[HTML]{FFF2CC} 
    \cellcolor[HTML]{FFF2CC} & Image & \textbf{99.15} & 98.99 & 98.76 & 96.37  \\
    \rowcolor[HTML]{FFF2CC} 
    \multirow{-2}{*}{\cellcolor[HTML]{FFF2CC}\textit{MC-AD}} & Pixel & \textbf{98.33} & 96.91 & 94.36  & 94.13 \\ \midrule
    \rowcolor[HTML]{D9D2E9} 
    \cellcolor[HTML]{D9D2E9}  & NMI & \textbf{0.5452} & 0.5287 & 0.5407 & 0.516 \\
    \rowcolor[HTML]{D9D2E9} 
    \cellcolor[HTML]{D9D2E9} & ARI  & \textbf{0.4331} & 0.4221 & 0.4266  & 0.3913 \\
    \rowcolor[HTML]{D9D2E9} 
    \multirow{-3}{*}{\cellcolor[HTML]{D9D2E9}\textit{Clustering}} & F1 & \textbf{0.6451} & 0.6357 & 0.6248 & 0.6129 \\ \bottomrule[1.0pt]
    \end{tabular}   
    }
}
\label{tab:bpm_thres}
\end{center}
\end{table}

\subsubsection{Comparison with Transformer-based Methods}\label{sec:vit_arch_comparison}
We compare our method with other transformer-based methods. In the case of PatchCore~\cite{roth2022towards}, we replaced ResNet with ViT-B/16 while keeping the other settings the same as in their original configuration. Table \ref{tab:vit_arch_comp} shows that our competitors exhibited poor performance in at least one scenario. In contrast, we consistently achieved robust performance across all scenarios, even when employing the same pre-trained encoder approach as PatchCore~\cite{roth2022towards}. Notably, PatchCore~\cite{roth2022towards} experienced a performance drop when transitioning to a ViT backbone, particularly evident in multiclass anomaly detection.  This highlights that the feature aggregation method utilized in the original paper~\cite{roth2022towards} is tailored to the ResNet backbone, also indicating that our performance is not solely attributable to the ViT architecture.

\begin{table}[ht!]
    \centering
    \caption{\textbf{Performance comparison with transformer-based competitors.} We report results on defect detection, semantic anomaly detection, and multi-class defect detection. Except for PatchCore~\cite{roth2022towards}, we borrowed results from ~\cite{lu2022unified}. For PatchCore, we replaced ResNet50 with ViT-B/16 and produced results ourselves with officially released code.}
    \resizebox{\linewidth}{!}{%
    \begin{tabular}{@{}cc|ccc|c@{}}
    \toprule[1.0pt]
                                                        &       & InTra~\cite{pirnay2022inpainting} & UniAD~\cite{lu2022unified} & PatchCore~\cite{roth2022towards} & Ours          \\ \midrule
    \rowcolor[HTML]{FFD5D5} 
    \cellcolor[HTML]{FFD5D5}                            & Image & 95.0  & \underline{96.6}  & 93.0            & \textbf{99.6} \\
    \rowcolor[HTML]{FFD5D5} 
    \multirow{-2}{*}{\cellcolor[HTML]{FFD5D5}\textit{Defect-AD}} & Pixel & 65.3  & \underline{96.5}  & 96.0            & \textbf{98.5} \\ \midrule
    \rowcolor[HTML]{DEF5F8} 
    \textit{Sem-AD}                                              & Image & -     & 52.9  & \underline{91.3}            & \textbf{92.6} \\ \midrule
    \rowcolor[HTML]{FFF2CC} 
    \cellcolor[HTML]{FFF2CC}                            & Image & \underline{96.6}  & \underline{96.6}  & 88.9            & \textbf{99.6} \\
    \rowcolor[HTML]{FFF2CC} 
    \multirow{-2}{*}{\cellcolor[HTML]{FFF2CC}\textit{MC-AD}}     & Pixel & 70.6  & \underline{96.8}  & 94.0            & \textbf{98.4} \\ \bottomrule[1.0pt]
    \end{tabular}
    }\label{tab:vit_arch_comp}
\end{table}


\section{Conclusion}
We propose a unified anomaly detection framework based on self-supervised representations after exploration of off-the-shelf representations suitable for unified anomaly detection. To achieve task unification in a task-agnostic manner, we introduce Back Patch Masking (BPM) and top $k$-ratio feature matching. Experimental results demonstrate that the proposed framework consistently achieves outstanding performance across various tasks, emphasizing the significance of BPM and top $k$-ratio within our framework. In this work, our primary focus is on visual anomaly detection, particularly in 2D images. However, in the real world, anomalies can emerge from various modalities, such as sensory data, video data, and natural language. Since the Vision Transformer (ViT) architecture is inherently scalable with respect to input modalities, there is potential for Uniformly to expand its task coverage to encompass other modalities. We intend to explore this avenue in future research.


\appendix

\begin{figure*}[hb!]
\begin{center}
    \includegraphics[width=\textwidth]{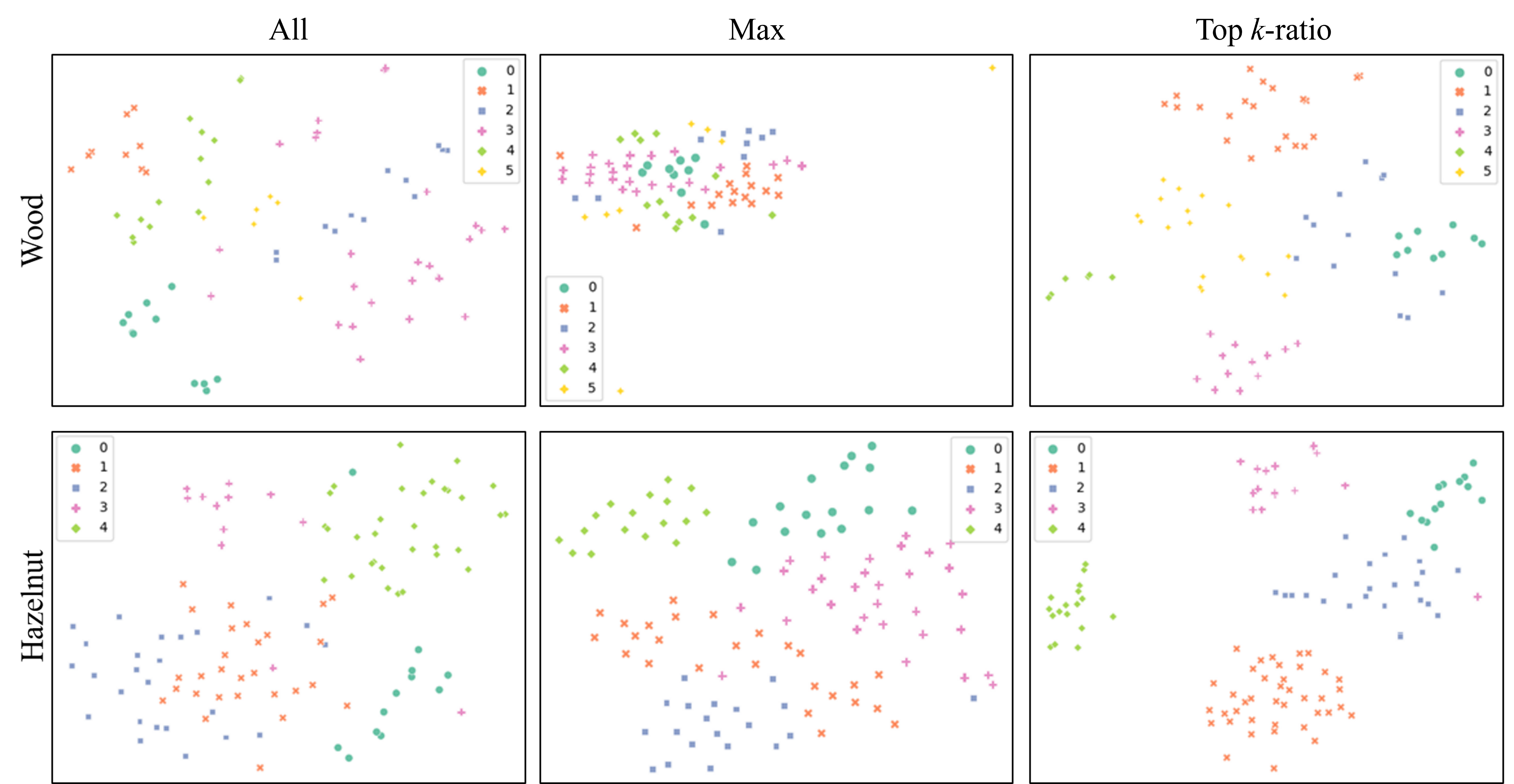}
\end{center}
\caption{\textbf{Visualization of each cluster grouped using \textit{UniFormaly} with different $k$.} We present t-SNE visualization of clustered anomalies within two classes in MVTecAD: the Wood (top) and Hazelnut (bottom). We visualize clusters using features with the maximum anomaly score (left), using all features (middle), and using features matched with the highest top $k$\% of anomaly scores (right).}
\label{fig:viz_clustering}
\end{figure*}

\section{Implementation Details}\label{sec:implementation_details}
We implemented our experiments on Python 3.8.10~\cite{10.5555/1593511} and PyTorch~\cite{NEURIPS2019_bdbca288} with NVIDIA A100 80GB GPUs on CUDA 11.3. For pretrained weights used in our method, we utilized the weights released on torchvision or their official repositories~\cite{caron2021emerging, zhou2021ibot} except for supervised ViTs. We used ImageNet-1K-pretrained weights of supervised ViTs-B/16 released in PyTorch Image Models~\cite{rw2019timm}. We preprocessed images by resizing to 256 $\times$ 256 and applying center crop to 224 $\times$ 224. For CIFAR datasets, we resized images to 112 $\times$ 112 pixels and added padding to reach 224 $\times$ 224 pixels. We conducted patch-level nearest neighbor retrieval using $\texttt{faiss}$~\cite{johnson2019billion}.


\section{Vision Transformers}\label{sec:vit}
We assume an input image $\mathrm{x} \in \mathbb{R}^{C \times H \times W}$, where $(H, W)$ is the resolution of image $\mathrm{x}$, and C is the number of channels. Vision Transformers~\cite{dosovitskiy2020image} requires a sequence of flattened, non-overlapping patches $\{\mathrm{x}^{(i)}| \mathrm{x}^{(i)} \in \mathbb{R}^{P^2 \cdot C}\}_{i=1}^N$ from the image $\mathrm{x}$, where $N$ is the number of patches and $(P, P)$ is the resolution of the patches. Each patch $x^{(i)}$ is then linearly projected into a $D$-dimensional vector $z \in \mathbb{R}^D$. The resulting patch embeddings are formulated as follows:

\begin{equation}
    \mathrm{z} = \{\mathrm{z}^{(i)} | \mathrm{z}^{(i)} \in \mathbb{R}^D \}_{i=1}^{N}.
\label{eq:patch_embedding}
\end{equation}

A learnable token $\mathrm{z}_{\texttt{[CLS]}}$ is prepended to the head of patch embeddings $\mathrm{z}$, which contains a global representation of an image. Consequently, ViTs process a sequence of embedded patches $\mathrm{z} = [\mathrm{z}_{\texttt{[CLS]}};\mathrm{z}^{(1)};\mathrm{z}^{(2)};...;\mathrm{z}^{(N-1)};\mathrm{z}^{(N)}]$ internally. Remark that the size of patch embeddings remained fixed during the entire encoding process as follows:

\begin{equation}
    \mathrm{z}_{l+1} = f_\theta^{(l)}(\mathrm{z}_l), \quad \mathrm{z}_l \in \mathbb{R}^{(N+1) \times D},
\end{equation}

where $l \in (1,L_{tot})$ and $L_{tot}$ is the total number of layers in the encoder $f_\theta$ parameterized with parameter $\theta$. Thus, $f_\theta^{(l)}$ indicates the $l$-th layer block of ViTs $f_\theta$.

\section{Species-60 Dataset}

\subsection{Details on Species-60 Dataset}
We randomly sampled 60 classes from Species datasets~\cite{hendrycks2022scaling}. There are about 500 training images per class and around 70 test images per class. The classes are present int Table \ref{tab:species_class}.

\subsection{Details on Split for Multi-Class Anomaly Detection}
For multi-class anomaly detection for Species-60, we construct 5 splits for evaluation. Each split consists of 30 randomly selected classes as normal. We provide normal class indices on each split in Table \ref{tab:multi_species_indices}. 

\section{Feature Space Visualization}\label{sec:feature_space_viz}

\subsection{Visualization of assigned clusters for anomalies}\label{subsec:topk_clustering}
In this section, we show top $k$-ratio feature matching is irreplaceable in anomaly clustering by comparing three available cases, All, Max, and Top $k$-ratio, regarding the preparation of features through feature visualization. For each case, we run clustering algorithms on the average pooled feature of all patch-level representations, the feature with the highest anomaly score, and the average pooled feature of patch features with the highest top $k$\% of anomaly scores.

\begin{figure*}[]
    \begin{center}
        \includegraphics[width=0.9\textwidth]{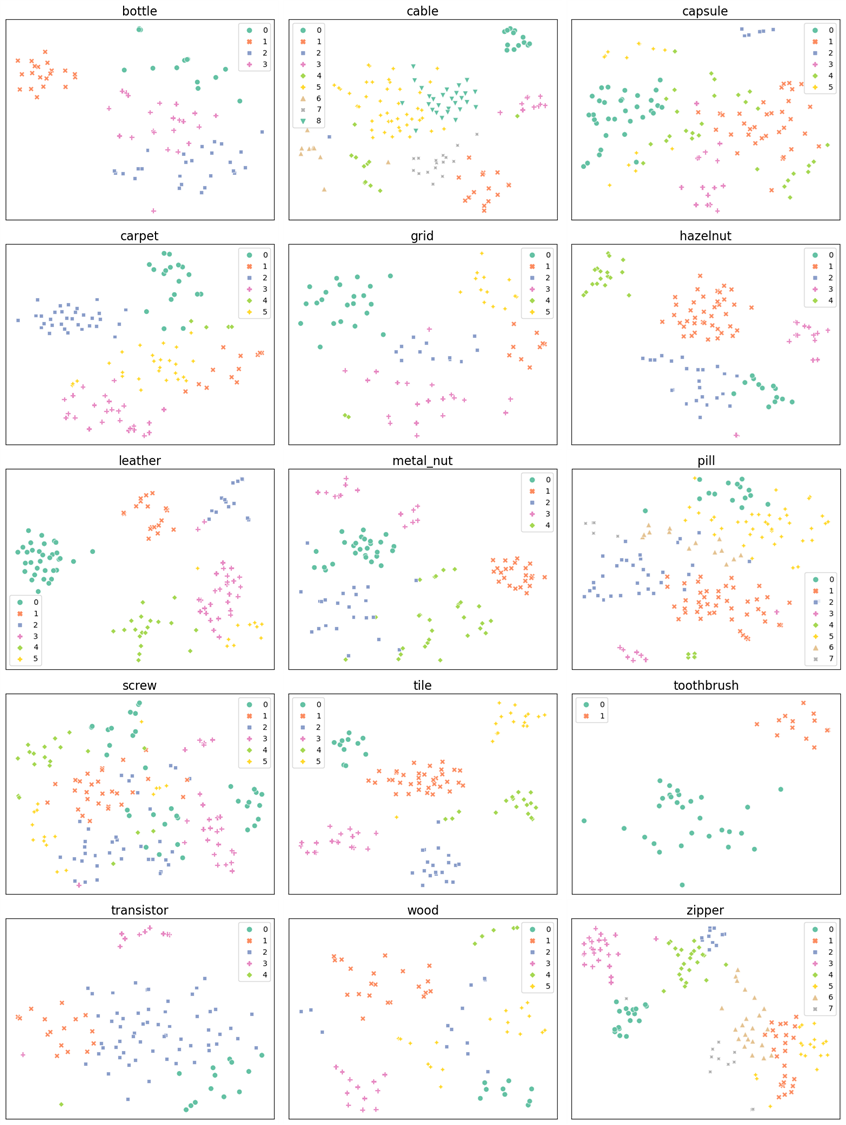}
        \caption{\textbf{Cluster visualization for all classes in MVTecAD.}}
        \label{fig:full_clustering_viz}
    \end{center}
\end{figure*}

The clustering results for two classes, Wood and Hazelnut, which belong to two supercategories in MVTecAD, objects and textures, were visually examined in Fig.~\ref{fig:viz_clustering}. It was observed that, for both classes and for the Max and All cases, the features were dispersed and did not form cohesive clusters. The resulting clusters were found to be impure compared to those obtained with the top $k$-ratio case. These qualitative findings are consistent with the quantitative results and provide additional support for the significance of top $k$-ratio feature matching in the context of anomaly clustering. We provide qualitative results for all classes in MVTecAD in Fig.~\ref{fig:full_clustering_viz}.

\begin{table*}[ht!]
\centering
\caption{\textbf{Classes in Species-60 Dataset.}}
\resizebox{1.0\linewidth}{!}{
\begin{tabular}{c|c|c|c|c|c}
\toprule[1.5pt]
\textbf{Index} & \textbf{Class Name} & \textbf{Index} & \textbf{Class Name} & \textbf{Index} & \textbf{Class Name} \\
\midrule
0 & abudefduf vaigiensis & 20 & halysidota tessellaris & 40 & leucocoprinus cepistipes \\
1 & acanthurus coeruleus & 21 & hippocampus whitei & 41 & lissachatina fulica \\
2 & acarospora socialis & 22 & hypomyces chrysospermus & 42 & lycium californicum \\
3 & acris gryllus & 23 & kuehneromyces mutabilis & 43 & megapallifera mutabilis \\
4 & adolphia clifornica & 24 & bjerkandera adusta & 44 & mononeuria groenlandica \\
5 & agaricus augustus & 25 & bombus pensylvanicus & 45 & neverita lewisii \\
6 & amanita parcivolvata & 26 & callianax biplicata & 46 & octopus tetricus \\
7 & ameiurus natalis & 27 & calochortus apiculatus & 47 & orienthella trilineata \\
8 & anartia jatrophae & 28 & campsis radicans & 48 & phyllotopsis nidulans \\
9 & aplysia californica & 29 & cepaea nemoralis & 49 & planorbarius corneus \\
10 & arcyria denudata & 30 & cladonia chlorophala & 50 & protoparmeliopsis muralis \\
11 & arion ater & 31 & cochlodina laminata & 51 & quercus dumosa \\
12 & aulostomus chinensis & 32 & crepidotus applanatus & 52 & ruditapes philippinarum \\
13 & bjerkandera adusta & 33 & cypripedium macranthos & 53 & salamandra lanzai \\
14 & bombus pensylvanicus & 34 & dendrobates auratus & 54 & swietenia macrophylla \\
15 & callianax biplicata & 35 & echinocereus pectinatus pectinatus & 55 & teloschistes chrysophthalmus \\
16 & calochortus apiculatus & 36 & eremnophila aureonotata & 56 & tramea onusta \\
17 & campsis radicans & 37 & etheostoma caeruleum & 57 & umbra limi \\
18 & cepaea nemoralis & 38 & fistularia commersonii & 58 & vespula squamosa \\
19 & cladonia chlorophala & 39 & ganoderma tsugae & 59 & zelus renardii \\
\bottomrule[1.5pt]
\end{tabular}
}
\label{tab:species_class}
\end{table*}

\begin{table*}[]
\centering
\caption{\textbf{Details on split for multi-class anomaly detection on Species-60.}}
\resizebox{1.0\textwidth}{!}{
\begin{tabular}{@{}c|c@{}}
\toprule
\textbf{Split} & \textbf{Classes}                                                                                                         \\ \midrule
A              & \{40, 21, 50, 36, 37, 2, 18, 44, 43, 22, 5, 46, 20, 26, 34, 11, 25, 59, 56, 16, 57, 42, 33, 19, 31, 54, 52, 10, 6, 38\}  \\ \midrule
B              & \{34, 26, 59, 40, 39, 54, 31, 50, 17, 21, 49, 55, 47, 36, 42, 6, 19, 30, 35, 44, 51, 37, 25, 53, 29, 7, 1, 14, 9, 48\}   \\ \midrule
C              & \{4, 51, 59, 47, 56, 16, 10, 55, 34, 36, 37, 39, 53, 32, 13, 21, 20, 15, 49, 58, 11, 12, 14, 17, 2, 35, 19, 30, 50, 38\} \\ \midrule
D              & \{19, 9, 30, 20, 3, 50, 15, 48, 41, 54, 25, 14, 22, 44, 38, 39, 2, 29, 21, 31, 24, 8, 52, 33, 35, 12, 57, 53, 43, 23\}   \\ \midrule
E              & \{22, 9, 41, 37, 11, 38, 24, 45, 18, 44, 27, 58, 35, 8, 46, 49, 54, 47, 23, 25, 1, 20, 50, 28, 26, 51, 16, 3, 42, 4\}    \\ \bottomrule
\end{tabular}
}
\label{tab:multi_species_indices}
\end{table*}

\section{How Back Patch Masking Works}

Back Patch Masking (BPM) aims to aid target-centric detection by isolating the foreground area containing the target(anomaly) from the background. However, it begs the question of the efficacy of BPM in scenarios where the input image primarily comprises a texture and lacks a distinct foreground object. Hence, we examine the distribution of the smoothed attention scores to observe whether BPM deteriorates to anomaly detection of non-object images. 

\begin{figure*}[]
    \begin{center}
        \includegraphics[width=1.0\textwidth]{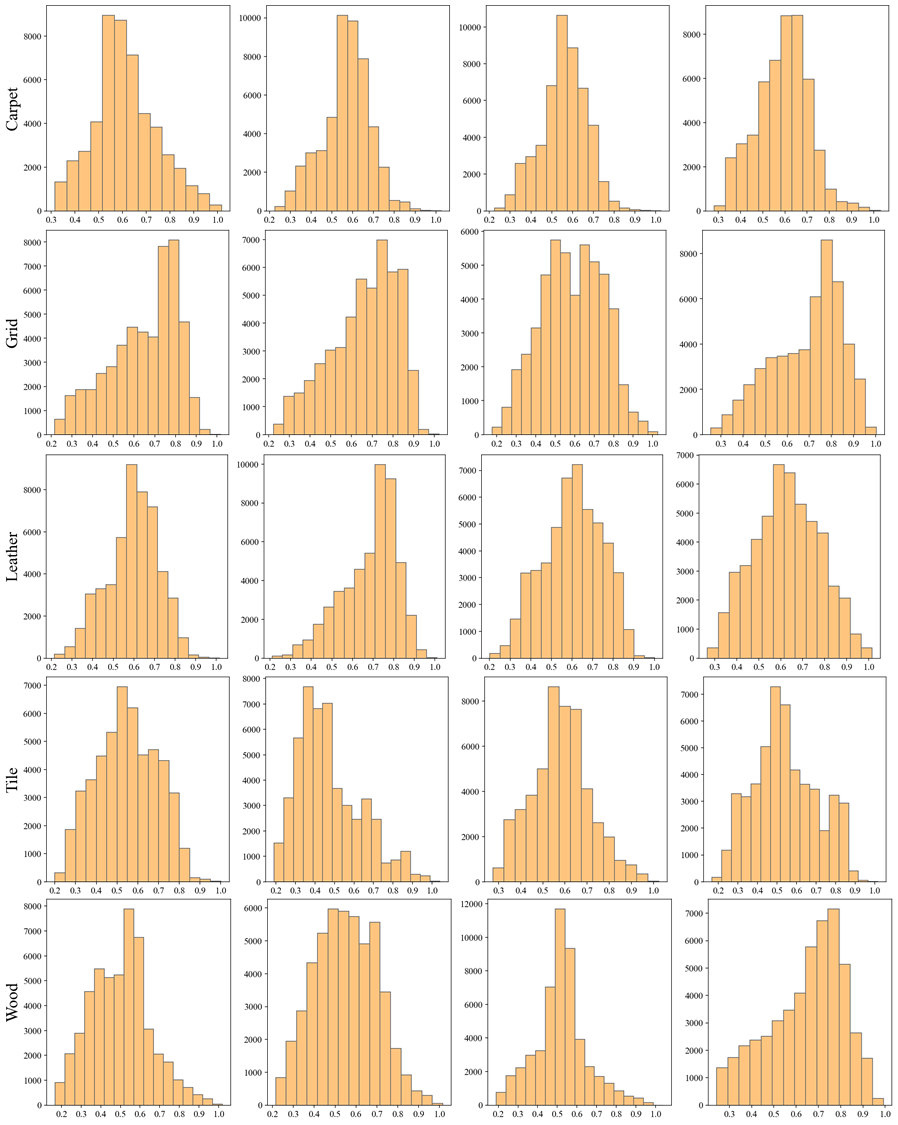}
        \caption{\textbf{Distribution of smoothed attention scores.} We randomly sampled four training images from each class in MVTecAD. For the classes of the objects category, the histogram is colored blue and orange for the classes from the textures.}
        \label{fig:attn_distn}
    \end{center}
\end{figure*}

Fig.\ref{fig:attn_distn} reveals that the distribution of attention scores in the textures category is shifted towards higher values, with the lowest attention score of approximately 0.2 compared to the objects category. Furthermore, within the objects category, the distribution of attention scores from several classes is highly skewed toward 0 to 0.1. In the case of the lowest attention score in the objects category, the value remains lower (0 vs. 0.2) than the lowest attention score observed in the textures category. This observation suggests that regions with attention values ranging from 0.1 to 1 in the objects category correspond to the background. Our default masking threshold of $\lambda$=0.1 did not yield any adverse effects on the images from the textures category or the images where objects are not present. It is important to note that we refrained from hyper-parameter tuning or manipulation on BPM with anomalous images from the textures category.

\begin{figure*}[]
    \begin{center}
        \includegraphics[width=1.0\textwidth]{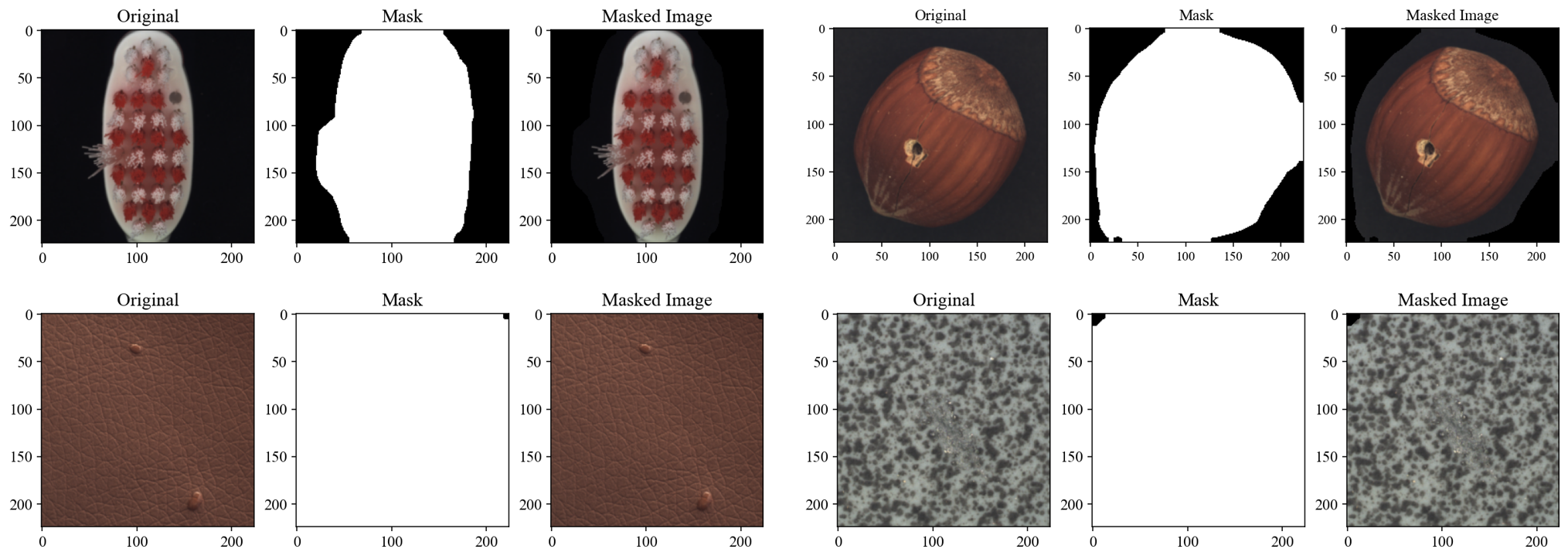}
        \includegraphics[width=1.0\textwidth]{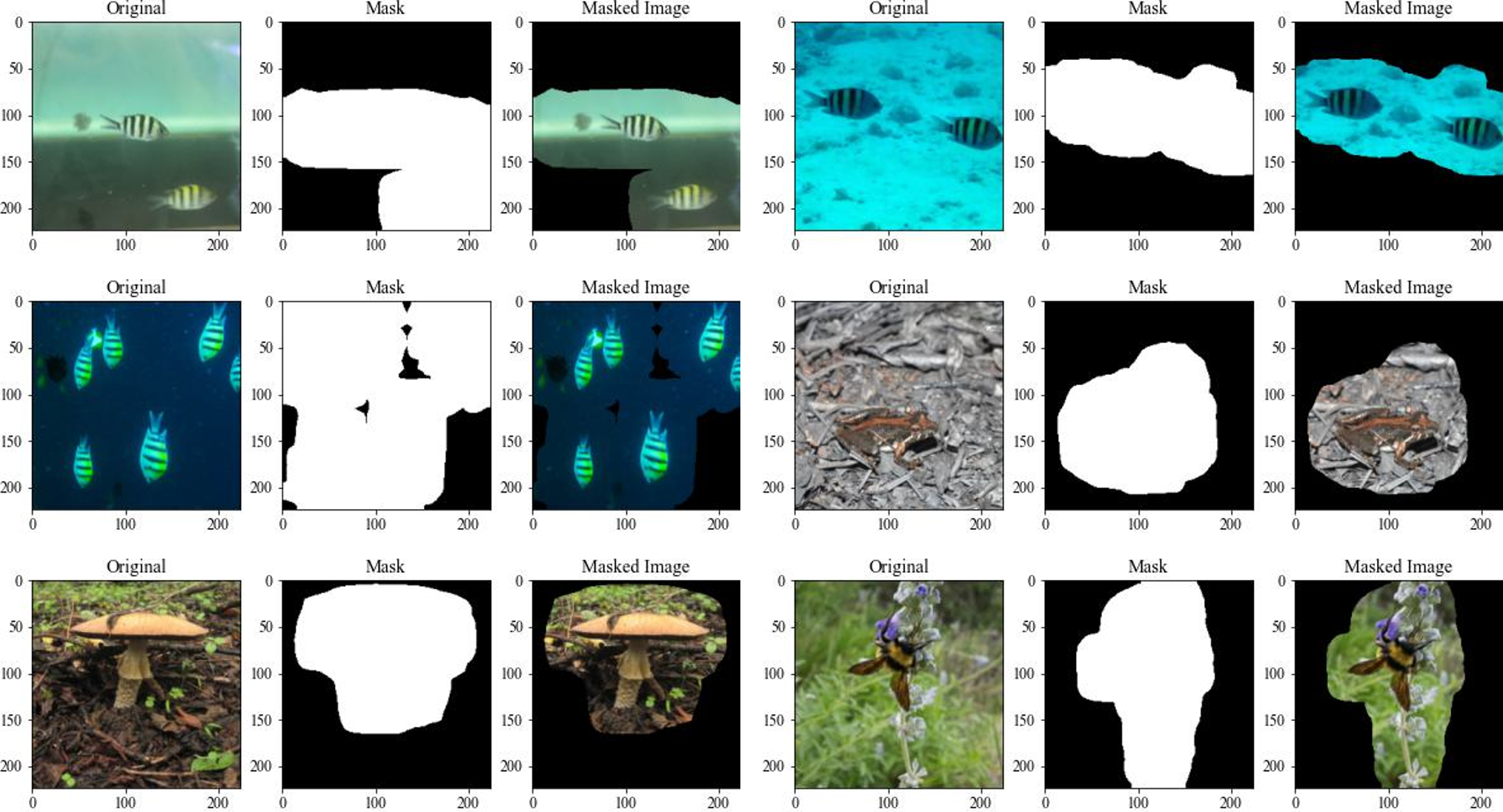}
        \caption{\textbf{Back Patch Masking visualization for object and texture classes in MVTecAD.} The first two rows present object and texture classes in MVTecAD, respectively. The last three rows visualize Back Patch Masking applied images from Species-60.}
        \label{fig:bpm_viz}
    \end{center}
\end{figure*}

We present images where BPM is applied in Figure \ref{fig:bpm_viz} to demonstrate that BPM is not activated in images without explicit objects. Furthermore, we observed that BPM effectively captures and distinguishes them from the background when multiple objects are present within a single image. This suggests that BPM is not limited to object-centric images alone; it has the potential to scale to tasks that require handling multiple classes or objects within a single image, such as anomaly segmentation in scene images.

\section{Full results for periodic evaluation of an online encoder-based method}\label{sec:periodic_evaluation}
\begin{figure*}[]
    \begin{center}
        \includegraphics[width=1.0\linewidth]
        {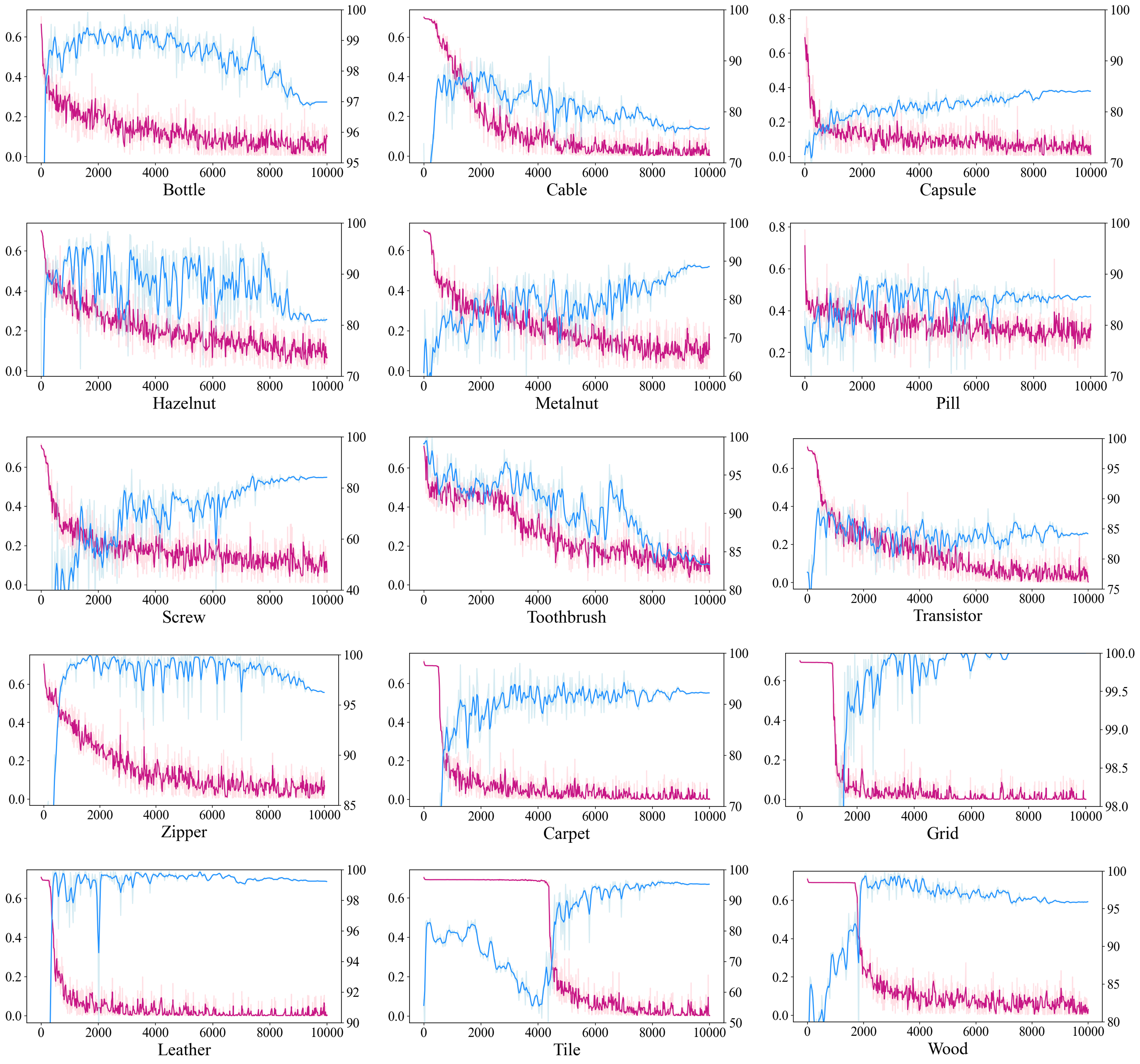}
        \caption{\textbf{Periodic evaluation of CutPaste~\cite{li2021cutpaste} on MVTecAD.} The blue line presents Image-level AUROC, and the red line plots the loss values for its proxy task.}
        \label{fig:periodic_evaluation_supp}
    \end{center}
\end{figure*}

In Fig.\ref{fig:periodic_evaluation_supp}, we show the full periodic evaluation results of self-supervised learning-based methods~\cite{li2021cutpaste} on the MVTecAD benchmark. As discussed earlier, we observed a discrepancy between its proxy and target tasks. 

\section{Full Results on Anomaly Detection Benchmark}
Tables \ref{tab:mvtec_full}, \ref{tab:cifar10_full}, \ref{tab:cifar100_full}, \ref{tab:species60_full},and \ref{tab:in30_full} display results for all classes within the MVTecAD, CIFAR-10, CIFAR-100, Species-60, and ImageNet-30 datasets, respectively.

\subsection{Visualization of memory bank for the multi-class case}
\begin{figure*}[]
    \begin{center}
        \includegraphics[width=0.9\linewidth]{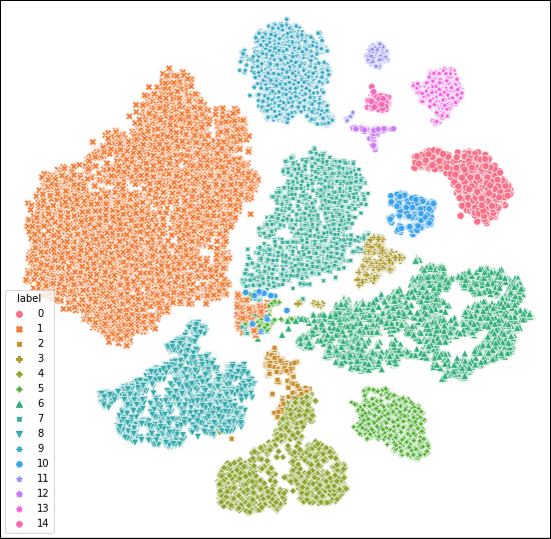}
    \end{center}
    \caption{\textbf{t-SNE~\cite{van2008visualizing} visualization for memory bank in multi-class anomaly detection.}}
    \label{fig:memory_bank}
\end{figure*}

We investigate the memory bank in the case of multi-class anomaly detection (Fig.\ref{fig:memory_bank}). We observe that each class is fairly separated from the other classes.


\begin{figure*}[]
\begin{center}
    \includegraphics[width=1.0\textwidth]{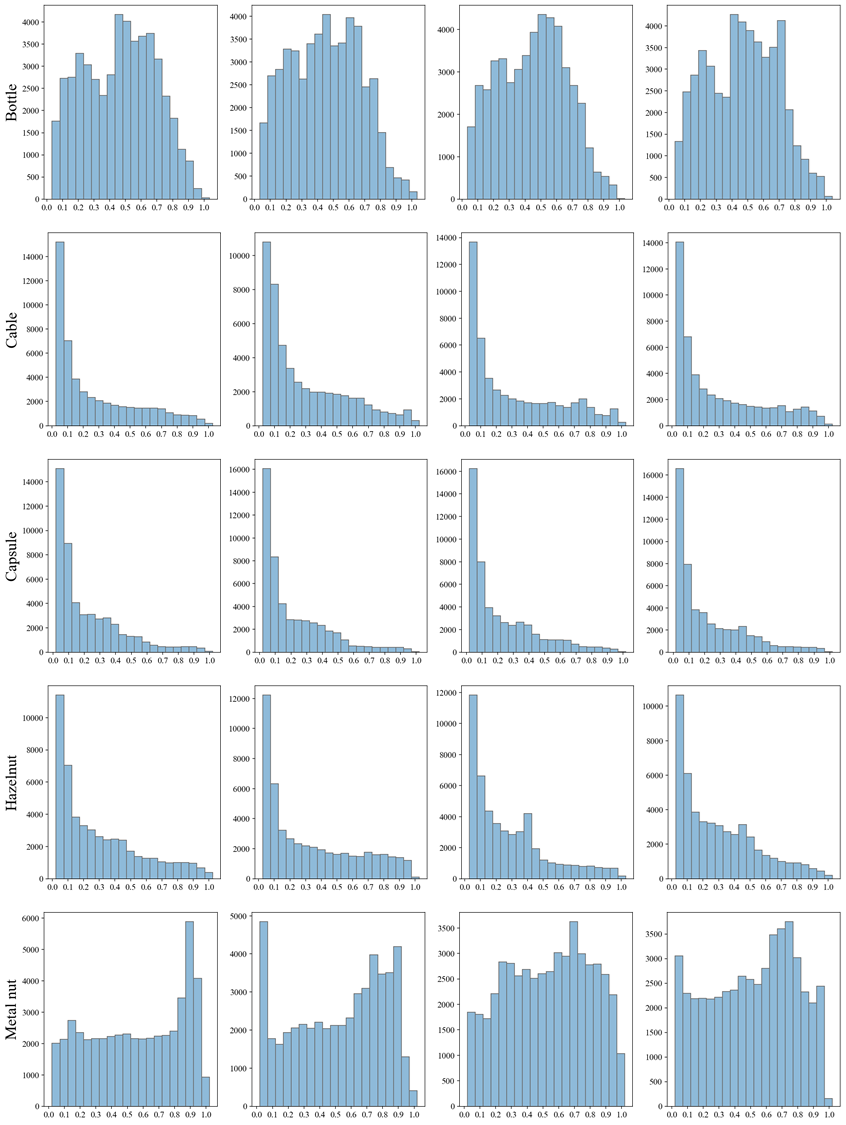}
\end{center}
\end{figure*}

\begin{figure*}[]
\begin{center}
    \includegraphics[width=1.0\textwidth]{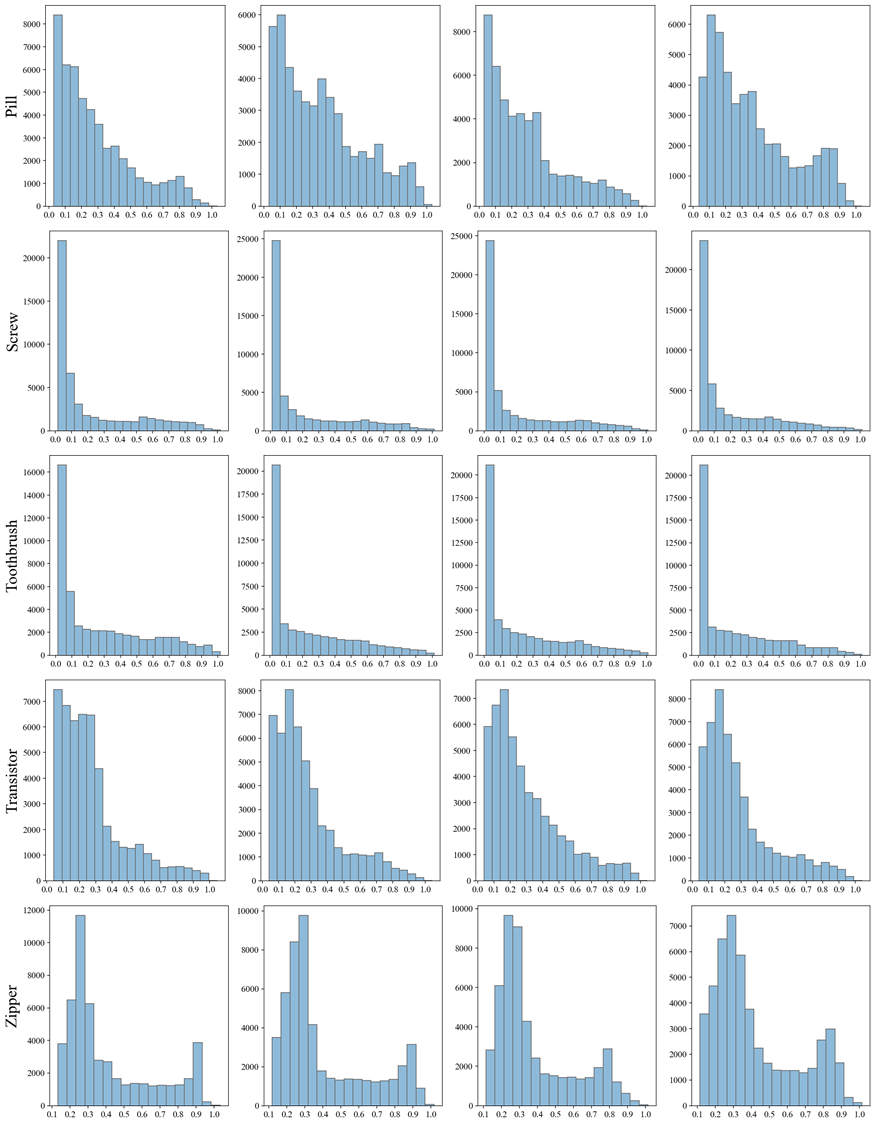}
\end{center}
\end{figure*}

\begin{table}[p]
\centering
\caption{\textbf{AUROC(\%) values of our method for MVTecAD.} Each row indicates the class results, and the last two rows indicate the mean and standard deviation values across 15 classes, respectively.}
\resizebox{\linewidth}{!}{
\begin{tabular}{@{}c|cc|cc@{}}
\toprule
\multirow{2}{*}{} & \multicolumn{2}{c|}{\textit{UniFormaly}-iBOT} & \multicolumn{2}{c}{\textit{UniFormaly}-DINO} \\ \cmidrule(l){2-5} 
                  & Image              & Pixel            & Image             & Pixel           \\ \midrule
screw             & 96.4               & 99.2             & 96.15             & 99.52           \\
pill              & 98.1               & 97.6             & 98.17             & 97.51           \\
capsule           & 97.6               & 98.9             & 98.88             & 99.03           \\
cable             & 99.8               & 98.3             & 100.00            & 97.67           \\
grid              & 97.9               & 98.7             & 98.50             & 99.37           \\
zipper            & 99.3               & 98.1             & 99.79             & 98.54           \\
transistor        & 99.0               & 97.3             & 99.25             & 96.45           \\
bottle            & 100.0              & 98.5             & 100.00            & 98.87           \\
carpet            & 99.4               & 99.1             & 99.56             & 99.38           \\
hazelnut          & 100.0              & 99.1             & 100.00            & 99.20           \\
leather           & 100.0              & 99.5             & 100.00            & 99.61           \\
metal nut         & 99.8               & 97.4             & 100.00            & 97.88           \\
tile              & 99.4               & 96.8             & 99.57             & 97.59           \\
toothbrush        & 100.0              & 98.7             & 100.00            & 98.86           \\
wood              & 100.0              & 97.2             & 99.91             & 97.62           \\ \midrule
Mean              &  99.11             & 98.3            & 99.32             & 98.47           \\
Std.               & 1.07               & 0.82             & 1.02              & 0.95            \\ \bottomrule
\end{tabular}}
\label{tab:mvtec_full}
\end{table}

\begin{table}[!]
\centering
\caption{\textbf{AUROC(\%) values of our method for CIFAR-10.} Each row indicates the results of the class, and the final row indicates the mean value.}
\resizebox{\linewidth}{!}{
\begin{tabular}{@{}c|cc@{}}
\toprule
Dataset          & \textit{UniFormaly}-iBOT & \textit{UniFormaly}-DINO \\ \midrule
airplane   & 95.7                     & 94.4                     \\
automobile & 97.9                     & 96.6                     \\
bird       & 90.6                     & 88.7                     \\
cat        & 87.1                     & 81.7                     \\
deer       & 96.0                     & 94.8                     \\
dog        & 91.0                     & 87.5                     \\
frog       & 97.8                     & 96.8                     \\
horse      & 96.6                     & 95.5                     \\
ship       & 97.4                     & 96.2                     \\
truck      & 97.3                     & 96.4                     \\ \midrule
Mean       & 94.7                     & 92.9            \\ \bottomrule
\end{tabular}
}
\label{tab:cifar10_full}
\end{table}

\begin{table}[ht!]
\centering
\caption{\textbf{AUROC(\%) values of our method for one-class CIFAR-100 (super-class).} Each row indicates the results of the selected super-class, and the final row indicates the mean value.}
\resizebox{0.8\linewidth}{!}{
\begin{tabular}{@{}c|cc@{}}
\toprule[0.7pt]
              & \textit{UniFormaly}-iBOT & \textit{UniFormaly}-DINO \\ \midrule
0             & 89.8            & 89.5            \\
1             & 91.0            & 89.5            \\
2             & 95.4            & 93.6            \\
3             & 85.1            & 85.4            \\
4             & 94.6            & 93.5            \\
5             & 86.5            & 86.4            \\
6             & 91.4            & 90.7            \\
7             & 89.7            & 86.2            \\
8             & 95.2            & 94.4            \\
9             & 91.8            & 90.8            \\
10            & 97.1            & 96.4            \\
11            & 88.0            & 86.4            \\
12            & 88.7            & 86.8            \\
13            & 87.7            & 84.8            \\
14            & 91.4            & 89.7            \\
15            & 85.6            & 85.7            \\
16            & 87.2            & 84.2            \\
17            & 98.7            & 97.9            \\
18            & 94.7            & 94.3            \\
19            & 93.2            & 93.0            \\ \midrule[0.7pt]
Mean          & 91.2            & 90.0   \\ \bottomrule[0.7pt]
\end{tabular}
}
\label{tab:cifar100_full}
\end{table}


\begin{table}[]
\centering
\caption{\textbf{AUROC(\%)values of our method for ImageNet-30.} Each row indicates the results of the class, and the final row indicates the mean value.}
\resizebox{0.8\linewidth}{!}{
\begin{tabular}{@{}c|cc@{}}
\toprule
                      & \textit{UniFormaly}-iBOT & \textit{UniFormaly}-DINO \\ \midrule
acorn                 & 99.1            & 99.1            \\
airliner              & 99.4            & 99.4            \\
ambulance             & 99.7            & 99.5            \\
american alligator    & 99.3            & 98.8            \\
banjo                 & 96.9            & 95.6            \\
barn                  & 98.9            & 98.6            \\
bikini                & 96.8            & 96.2            \\
digital clock         & 96.1            & 95.5            \\
dragonfly             & 98.9            & 98.8            \\
dumbbell              & 94.8            & 94.4            \\
forklift              & 98.5            & 98.5            \\
goblet                & 94.9            & 90.4            \\
grand piano           & 98.9            & 97.3            \\
hotdog                & 95.5            & 94.2            \\
hourglass             & 97.1            & 95.8            \\
manhole cover         & 99.9            & 99.9            \\
mosque                & 99.1            & 98.8            \\
nail                  & 97.4            & 97.2            \\
parking meter         & 94.8            & 93.2            \\
pillow                & 96.0            & 94.9            \\
revolver              & 96.3            & 95.6            \\
rotary dial telephone & 98.8            & 98.1            \\
schooner              & 98.0            & 97.5            \\
snowmobile            & 99.2            & 99.1            \\
soccer ball           & 93.3            & 92.5            \\
stingray              & 99.4            & 98.5            \\
strawberry            & 98.7            & 98.4            \\
tank                  & 99.0            & 98.2            \\
toaster               & 95.1            & 92.1            \\
volcano               & 99.3            & 98.8            \\ \midrule
Mean                  & 97.6            & 96.8            \\ \bottomrule
\end{tabular}
}
\label{tab:in30_full}
\end{table}

\begin{table*}[]
\begin{center}
\caption{\textbf{AUROC(\%) values of our method for one-class Species-60.} Each column indicates the results of the class. iBOT and DINO represent \textit{UniFormaly}-iBOT and \textit{UniFormaly}-DINO, respectively.}
\resizebox{1.0\textwidth}{!}{
\begin{tabular}{@{}c|ccccccccccccccc@{}}
\toprule[1.0pt]
     & 0    & 1    & 2    & 3    & 4    & 5    & 6    & 7    & 8    & 9    & 10   & 11   & 12   & 13   & 14   \\ \midrule
iBOT & 96.2 & 98.1 & 98.1 & 92.0 & 98.1 & 89.0 & 94.3 & 85.9 & 96.8 & 90.6 & 95.8 & 95.8 & 96.1 & 92.5 & 91.0 \\ 
DINO & 94.1 & 97.0 & 98.0 & 89.1 & 98.0 & 86.3 & 94.6 & 81.2 & 96.1 & 87.2 & 95.0 & 92.0 & 94.3 & 91.9 & 88.3 \\
\midrule\midrule
     & 15   & 16   & 17   & 18   & 19   & 20   & 21   & 22   & 23   & 24   & 25   & 26   & 27   & 28   & 29   \\ \midrule
iBOT & 95.3 & 99.1 & 92.4 & 88.4 & 95.3 & 92.1 & 93.9 & 99.4 & 94.5 & 97.5 & 97.1 & 95.7 & 95.6 & 96.6 & 94.9 \\
DINO & 91.3 & 98.8 & 92.2 & 86.0 & 95.5 & 90.1 & 93.4 & 99.4 & 92.1 & 95.0 & 96.1 & 96.1 & 94.0 & 94.6 & 94.4 \\
 \midrule\midrule
     & 30   & 31   & 32   & 33   & 34   & 35   & 36   & 37   & 38   & 39   & 40   & 41   & 42   & 43   & 44   \\ \midrule
iBOT & 92.2 & 87.9 & 91.2 & 98.2 & 88.9 & 92.1 & 95.9 & 95.3 & 95.8 & 91.9 & 90.7 & 88.2 & 97.5 & 91.6 & 98.2 \\
DINO & 89.5 & 86.3 & 87.3 & 98.2 & 89.0 & 92.1 & 95.3 & 94.3 & 94.3 & 90.3 & 90.2 & 86.7 & 96.6 & 90.1 & 98.3 \\
 \midrule\midrule
     & 45   & 46   & 47   & 48   & 49   & 50   & 51   & 52   & 53   & 54   & 55   & 56   & 57   & 58   & 59   \\ \midrule
iBOT & 92.9 & 95.4 & 98.7 & 92.2 & 88.6 & 99.2 & 98.0 & 92.4 & 97.1 & 90.8 & 96.1 & 99.2 & 94.2 & 91.3 & 92.7 \\ 
DINO & 86.9 & 93.9 & 96.8 & 91.2 & 85.7 & 99.2 & 97.4 & 89.6 & 97.4 & 91.6 & 93.1 & 98.3 & 90.9 & 86.3 & 89.4 \\
\bottomrule[1.0pt]
\end{tabular}}
\label{tab:species60_full}
\end{center}
\end{table*}


{\small
\bibliographystyle{ieee_fullname}
\bibliography{egbib}

\begin{thebibliography}{10}\itemsep=-1pt

\bibitem{ahn2022application}
Jae-Young Ahn and Gyeonghwan Kim.
\newblock Application of optimal clustering and metric learning to patch-based anomaly detection.
\newblock {\em Pattern Recognition Letters}, 154:110--115, 2022.

\bibitem{bergman2020deep}
Liron Bergman, Niv Cohen, and Yedid Hoshen.
\newblock Deep nearest neighbor anomaly detection.
\newblock {\em arXiv preprint arXiv:2002.10445}, 2020.

\bibitem{bergmann2019mvtec}
Paul Bergmann, Michael Fauser, David Sattlegger, and Carsten Steger.
\newblock Mvtec ad--a comprehensive real-world dataset for unsupervised anomaly detection.
\newblock In {\em Proceedings of the IEEE/CVF conference on computer vision and pattern recognition}, pages 9592--9600, 2019.

\bibitem{bergmann2020uninformed}
Paul Bergmann, Michael Fauser, David Sattlegger, and Carsten Steger.
\newblock Uninformed students: Student-teacher anomaly detection with discriminative latent embeddings.
\newblock In {\em Proceedings of the IEEE/CVF conference on computer vision and pattern recognition}, pages 4183--4192, 2020.

\bibitem{caron2020unsupervised}
Mathilde Caron, Ishan Misra, Julien Mairal, Priya Goyal, Piotr Bojanowski, and Armand Joulin.
\newblock Unsupervised learning of visual features by contrasting cluster assignments.
\newblock {\em Advances in neural information processing systems}, 33:9912--9924, 2020.

\bibitem{caron2021emerging}
Mathilde Caron, Hugo Touvron, Ishan Misra, Herv{\'e} J{\'e}gou, Julien Mairal, Piotr Bojanowski, and Armand Joulin.
\newblock Emerging properties in self-supervised vision transformers.
\newblock In {\em Proceedings of the IEEE/CVF International Conference on Computer Vision}, pages 9650--9660, 2021.

\bibitem{cohen2020sub}
Niv Cohen and Yedid Hoshen.
\newblock Sub-image anomaly detection with deep pyramid correspondences.
\newblock {\em arXiv preprint arXiv:2005.02357}, 2020.

\bibitem{defard2021padim}
Thomas Defard, Aleksandr Setkov, Angelique Loesch, and Romaric Audigier.
\newblock Padim: a patch distribution modeling framework for anomaly detection and localization.
\newblock In {\em International Conference on Pattern Recognition}, pages 475--489. Springer, 2021.

\bibitem{ding2022catching}
Choubo Ding, Guansong Pang, and Chunhua Shen.
\newblock Catching both gray and black swans: Open-set supervised anomaly detection.
\newblock In {\em Proceedings of the IEEE/CVF Conference on Computer Vision and Pattern Recognition}, pages 7388--7398, 2022.

\bibitem{dosovitskiy2020image}
Alexey Dosovitskiy, Lucas Beyer, Alexander Kolesnikov, Dirk Weissenborn, Xiaohua Zhai, Thomas Unterthiner, Mostafa Dehghani, Matthias Minderer, Georg Heigold, Sylvain Gelly, et~al.
\newblock An image is worth 16x16 words: Transformers for image recognition at scale.
\newblock {\em arXiv preprint arXiv:2010.11929}, 2020.

\bibitem{he2016deep}
Kaiming He, Xiangyu Zhang, Shaoqing Ren, and Jian Sun.
\newblock Deep residual learning for image recognition.
\newblock In {\em Proceedings of the IEEE conference on computer vision and pattern recognition}, pages 770--778, 2016.

\bibitem{hendrycks2022scaling}
Dan Hendrycks, Steven Basart, Mantas Mazeika, Andy Zou, Joseph Kwon, Mohammadreza Mostajabi, Jacob Steinhardt, and Dawn Song.
\newblock Scaling out-of-distribution detection for real-world settings.
\newblock In {\em International Conference on Machine Learning}, pages 8759--8773. PMLR, 2022.

\bibitem{hendrycks2019using}
Dan Hendrycks, Mantas Mazeika, Saurav Kadavath, and Dawn Song.
\newblock Using self-supervised learning can improve model robustness and uncertainty.
\newblock {\em Advances in neural information processing systems}, 32, 2019.

\bibitem{huang2020surface}
Yibin Huang, Congying Qiu, and Kui Yuan.
\newblock Surface defect saliency of magnetic tile.
\newblock {\em The Visual Computer}, 36:85--96, 2020.

\bibitem{ji2019invariant}
Xu Ji, Joao~F Henriques, and Andrea Vedaldi.
\newblock Invariant information clustering for unsupervised image classification and segmentation.
\newblock In {\em Proceedings of the IEEE/CVF International Conference on Computer Vision}, pages 9865--9874, 2019.

\bibitem{johnson2019billion}
Jeff Johnson, Matthijs Douze, and Herv{\'e} J{\'e}gou.
\newblock Billion-scale similarity search with {GPUs}.
\newblock {\em IEEE Transactions on Big Data}, 7(3):535--547, 2019.

\bibitem{krizhevsky2009learning}
Alex Krizhevsky and Geoffrey Hinton.
\newblock Learning multiple layers of features from tiny images.
\newblock Technical Report~0, University of Toronto, Toronto, Ontario, 2009.

\bibitem{kuhn1955hungarian}
Harold~W Kuhn.
\newblock The hungarian method for the assignment problem.
\newblock {\em Naval research logistics quarterly}, 2(1-2):83--97, 1955.

\bibitem{li2021cutpaste}
Chun-Liang Li, Kihyuk Sohn, Jinsung Yoon, and Tomas Pfister.
\newblock Cutpaste: Self-supervised learning for anomaly detection and localization.
\newblock In {\em Proceedings of the IEEE/CVF Conference on Computer Vision and Pattern Recognition}, pages 9664--9674, 2021.

\bibitem{liznerski2021explainable}
Philipp Liznerski, Lukas Ruff, Robert~A. Vandermeulen, Billy~Joe Franks, Marius Kloft, and Klaus~Robert Muller.
\newblock Explainable deep one-class classification.
\newblock In {\em International Conference on Learning Representations}, 2021.

\bibitem{lu2022unified}
Jiasen Lu, Christopher Clark, Rowan Zellers, Roozbeh Mottaghi, and Aniruddha Kembhavi.
\newblock Unified-io: A unified model for vision, language, and multi-modal tasks.
\newblock {\em arXiv preprint arXiv:2206.08916}, 2022.

\bibitem{mishra2021vt}
Pankaj Mishra, Riccardo Verk, Daniele Fornasier, Claudio Piciarelli, and Gian~Luca Foresti.
\newblock Vt-adl: A vision transformer network for image anomaly detection and localization.
\newblock In {\em 2021 IEEE 30th International Symposium on Industrial Electronics (ISIE)}, pages 01--06. IEEE, 2021.

\bibitem{niu2020gatcluster}
Chuang Niu, Jun Zhang, Ge Wang, and Jimin Liang.
\newblock Gatcluster: Self-supervised gaussian-attention network for image clustering.
\newblock In {\em Computer Vision--ECCV 2020: 16th European Conference, Glasgow, UK, August 23--28, 2020, Proceedings, Part XXV 16}, pages 735--751. Springer, 2020.

\bibitem{NEURIPS2019_bdbca288}
Adam Paszke, Sam Gross, Francisco Massa, Adam Lerer, James Bradbury, Gregory Chanan, Trevor Killeen, Zeming Lin, Natalia Gimelshein, Luca Antiga, Alban Desmaison, Andreas Kopf, Edward Yang, Zachary DeVito, Martin Raison, Alykhan Tejani, Sasank Chilamkurthy, Benoit Steiner, Lu Fang, Junjie Bai, and Soumith Chintala.
\newblock Pytorch: An imperative style, high-performance deep learning library.
\newblock In H. Wallach, H. Larochelle, A. Beygelzimer, F. d\textquotesingle Alch\'{e}-Buc, E. Fox, and R. Garnett, editors, {\em Advances in Neural Information Processing Systems}, volume~32. Curran Associates, Inc., 2019.

\bibitem{pirnay2022inpainting}
Jonathan Pirnay and Keng Chai.
\newblock Inpainting transformer for anomaly detection.
\newblock In {\em International Conference on Image Analysis and Processing}, pages 394--406. Springer, 2022.

\bibitem{reiss2021panda}
Tal Reiss, Niv Cohen, Liron Bergman, and Yedid Hoshen.
\newblock Panda: Adapting pretrained features for anomaly detection and segmentation.
\newblock In {\em Proceedings of the IEEE/CVF Conference on Computer Vision and Pattern Recognition}, pages 2806--2814, 2021.

\bibitem{roth2022towards}
Karsten Roth, Latha Pemula, Joaquin Zepeda, Bernhard Sch{\"o}lkopf, Thomas Brox, and Peter Gehler.
\newblock Towards total recall in industrial anomaly detection.
\newblock In {\em Proceedings of the IEEE/CVF Conference on Computer Vision and Pattern Recognition}, pages 14318--14328, 2022.

\bibitem{sohn2023anomaly}
Kihyuk Sohn, Jinsung Yoon, Chun-Liang Li, Chen-Yu Lee, and Tomas Pfister.
\newblock Anomaly clustering: Grouping images into coherent clusters of anomaly types.
\newblock In {\em Proceedings of the IEEE/CVF Winter Conference on Applications of Computer Vision}, pages 5479--5490, 2023.

\bibitem{sun2022out}
Yiyou Sun, Yifei Ming, Xiaojin Zhu, and Yixuan Li.
\newblock Out-of-distribution detection with deep nearest neighbors.
\newblock In {\em International Conference on Machine Learning}, pages 20827--20840. PMLR, 2022.

\bibitem{tack2020csi}
Jihoon Tack, Sangwoo Mo, Jongheon Jeong, and Jinwoo Shin.
\newblock Csi: Novelty detection via contrastive learning on distributionally shifted instances.
\newblock {\em Advances in neural information processing systems}, 33:11839--11852, 2020.

\bibitem{van2008visualizing}
Laurens Van~der Maaten and Geoffrey Hinton.
\newblock Visualizing data using t-sne.
\newblock {\em Journal of machine learning research}, 9(11), 2008.

\bibitem{van2020scan}
Wouter Van~Gansbeke, Simon Vandenhende, Stamatios Georgoulis, Marc Proesmans, and Luc Van~Gool.
\newblock Scan: Learning to classify images without labels.
\newblock In {\em Computer Vision--ECCV 2020: 16th European Conference, Glasgow, UK, August 23--28, 2020, Proceedings, Part X}, pages 268--285. Springer, 2020.

\bibitem{10.5555/1593511}
Guido Van~Rossum and Fred~L. Drake.
\newblock {\em Python 3 Reference Manual}.
\newblock CreateSpace, Scotts Valley, CA, 2009.

\bibitem{venkataramanan2020attention}
Shashanka Venkataramanan, Kuan-Chuan Peng, Rajat~Vikram Singh, and Abhijit Mahalanobis.
\newblock Attention guided anomaly localization in images.
\newblock In {\em European Conference on Computer Vision}, pages 485--503. Springer, 2020.

\bibitem{rw2019timm}
Ross Wightman.
\newblock Pytorch image models.
\newblock \url{https://github.com/rwightman/pytorch-image-models}, 2019.

\bibitem{wu2021learning}
Jhih-Ciang Wu, Ding-Jie Chen, Chiou-Shann Fuh, and Tyng-Luh Liu.
\newblock Learning unsupervised metaformer for anomaly detection.
\newblock In {\em Proceedings of the IEEE/CVF International Conference on Computer Vision}, pages 4369--4378, 2021.

\bibitem{yang2022learning}
Jie Yang, Yong Shi, and Zhiquan Qi.
\newblock Learning deep feature correspondence for unsupervised anomaly detection and segmentation.
\newblock {\em Pattern Recognition}, 132:108874, 2022.

\bibitem{yi2020patch}
Jihun Yi and Sungroh Yoon.
\newblock Patch svdd: Patch-level svdd for anomaly detection and segmentation.
\newblock In {\em Proceedings of the Asian Conference on Computer Vision}, 2020.

\bibitem{you2022a}
Zhiyuan You, Lei Cui, Yujun Shen, Kai Yang, Xin Lu, Yu Zheng, and Xinyi Le.
\newblock A unified model for multi-class anomaly detection.
\newblock In Alice~H. Oh, Alekh Agarwal, Danielle Belgrave, and Kyunghyun Cho, editors, {\em Advances in Neural Information Processing Systems}, 2022.

\bibitem{zavrtanik2021draem}
Vitjan Zavrtanik, Matej Kristan, and Danijel Sko{\v{c}}aj.
\newblock Draem-a discriminatively trained reconstruction embedding for surface anomaly detection.
\newblock In {\em Proceedings of the IEEE/CVF International Conference on Computer Vision}, pages 8330--8339, 2021.

\bibitem{zavrtanik2021reconstruction}
Vitjan Zavrtanik, Matej Kristan, and Danijel Sko{\v{c}}aj.
\newblock Reconstruction by inpainting for visual anomaly detection.
\newblock {\em Pattern Recognition}, 112:107706, 2021.

\bibitem{zhang2022pedenet}
Kaitai Zhang, Bin Wang, and C-C~Jay Kuo.
\newblock Pedenet: Image anomaly localization via patch embedding and density estimation.
\newblock {\em Pattern Recognition Letters}, 153:144--150, 2022.

\bibitem{zhang2022deep}
Xianchao Zhang, Jie Mu, Xiaotong Zhang, Han Liu, Linlin Zong, and Yuangang Li.
\newblock Deep anomaly detection with self-supervised learning and adversarial training.
\newblock {\em Pattern Recognition}, 121:108234, 2022.

\bibitem{zhang2020multi}
Yingying Zhang, Yuxin Gong, Haogang Zhu, Xiao Bai, and Wenzhong Tang.
\newblock Multi-head enhanced self-attention network for novelty detection.
\newblock {\em Pattern Recognition}, 107:107486, 2020.

\bibitem{zhou2021ibot}
Jinghao Zhou, Chen Wei, Huiyu Wang, Wei Shen, Cihang Xie, Alan Yuille, and Tao Kong.
\newblock ibot: Image bert pre-training with online tokenizer.
\newblock {\em arXiv preprint arXiv:2111.07832}, 2021.

\end{thebibliography}
}

\end{document}